\documentclass[10pt,journal,compsoc]{IEEEtran}

\usepackage{ifpdf}

\ifCLASSOPTIONcompsoc
  \usepackage[nocompress]{cite}
\else
  \usepackage{cite}
\fi

\ifCLASSINFOpdf
 
\else
  
\fi

\usepackage{amsmath}
\usepackage{algorithmic}
\usepackage{array}
\usepackage{stfloats}
\usepackage{url}
\usepackage{bbding}

\usepackage{algorithm}

\makeatletter
\let\NAT@parse\undefined
\makeatother
\usepackage{overpic}
\usepackage{amsfonts,amssymb}
\usepackage{subfig}
\usepackage{booktabs}
\usepackage{makecell}
\usepackage{enumitem}
\usepackage{tikz}
\usepackage{forest}
\usepackage{tabularx}
\usepackage{caption}
\usepackage[square,numbers]{natbib} 
\usepackage{multicol}
\usepackage{colortbl} 
\usepackage{adjustbox}
\usetikzlibrary{shapes.geometric, positioning, arrows.meta, shadows}
\usetikzlibrary{mindmap, trees}
\definecolor{st_color}{RGB}{0, 153, 230}
\definecolor{t_color}{RGB}{255, 102, 102}

\definecolor{hidden-draw}{RGB}{20,68,106}
\definecolor{hidden-pink}{RGB}{255,245,247}
\usepackage{bm}

\usepackage{graphicx}
\usepackage{multirow}
\usepackage{xurl}
\usepackage{wrapfig,lipsum}
\usepackage{float}
\usepackage[numbers]{natbib}
\usepackage{xcolor}
\usepackage{pifont}
\usepackage{verbatim}

\usepackage{hyperref}

\begin{document}

\title{A Survey on Video Temporal Grounding with Multimodal Large Language Model}

\author{
    Jianlong~Wu,~\IEEEmembership{Member,~IEEE,}
    Wei~Liu,
    Ye~Liu,
    Meng~Liu,~\IEEEmembership{Member,~IEEE,}\\
    Liqiang~Nie,~\IEEEmembership{Senior Member,~IEEE,}
    Zhouchen~Lin,~\IEEEmembership{Fellow,~IEEE,}
    and Chang~Wen~Chen,~\IEEEmembership{Fellow,~IEEE}
    \thanks{$\bullet$ Jianlong Wu, Wei Liu, and Liqiang Nie are with the School of Computer Science and Technology, Harbin Institute of Technology, Shenzhen 518055, China (e-mail: wujianlong@hit.edu.cn; liuwei030224@gmail.com; nieliqiang@gmail.com).}
    \thanks{$\bullet$ Meng Liu is with the School of Computer Science and Technology, Shandong Jianzhu University, Jinan 250101, China (e-mail: mengliu.sdu@gmail.com).}
    \thanks{$\bullet$ Zhouchen Lin is with the School of Intelligence Science and Technology, Peking University, Beijing 100871, China (e-mail: zlin@pku.edu.cn).}
    \thanks{$\bullet$ Ye Liu and Chang Wen Chen are with the Department of Computing, The Hong Kong Polytechnic University, Hong Kong, China (e-mail: coco.ye.liu@connect.polyu.hk; changwen.chen@polyu.edu.hk).}
}

\IEEEtitleabstractindextext{
\begin{abstract}
The recent advancement in video temporal grounding (VTG) has significantly enhanced fine-grained video understanding, primarily driven by multimodal large language models (MLLMs). With superior multimodal comprehension and reasoning abilities, VTG approaches based on MLLMs (VTG-MLLMs) are gradually surpassing traditional fine-tuned methods. They not only achieve competitive performance but also excel in generalization across zero-shot, multi-task, and multi-domain settings. Despite extensive surveys on general video-language understanding, comprehensive reviews specifically addressing VTG-MLLMs remain scarce. To fill this gap, this survey systematically examines current research on VTG-MLLMs through a three-dimensional taxonomy: 1) the functional roles of MLLMs, highlighting their architectural significance; 2) training paradigms, analyzing strategies for temporal reasoning and task adaptation; and 3) video feature processing techniques, which determine spatiotemporal representation effectiveness. We further discuss benchmark datasets, evaluation protocols, and summarize empirical findings. Finally, we identify existing limitations and propose promising research directions. For additional resources and details, readers are encouraged to visit our repository at \url{https://github.com/ki-lw/Awesome-MLLMs-for-Video-Temporal-Grounding}.
\end{abstract}

}

% make the title area
\maketitle

\IEEEdisplaynontitleabstractindextext
\IEEEpeerreviewmaketitle
\IEEEraisesectionheading{\section{Introduction}\label{sec:introduction}}

\label{introduction}
\IEEEPARstart{T}{he} proliferation of untrimmed video content across domains such as surveillance, entertainment, and autonomous systems has created an urgent need for systems capable of precise temporal understanding. Real-world applications, including moment retrieval, scene editing, and temporal question answering, demand accurate identification of not only what events occur but precisely when they occur.  Existing video-language models primarily focus on global or coarse-level video comprehension~\cite{wang2022internvideo,wang2024internvideo2,wang2023actionclip,yang2023vid2seq},  making them inadequate for tasks requiring fine-grained temporal grounding of events described by natural language. To address this capability gap, \textit{Video Temporal Grounding} (VTG) has emerged as a pivotal research area. VTG involves localizing video segments that correspond specifically to given textual queries, enabling detailed interaction with video content. The core challenge of VTG lies in precisely aligning complex linguistic semantics with temporally distributed visual information, while simultaneously handling complex temporal relationships within the video.

As illustrated in Fig.~\ref{Fig:Illustration-tasks}, VTG encompasses several closely related but distinct tasks: (a) \textit{Video Moment Retrieval}~\cite{regneri2013grounding, anne2017localizing}, where the goal is to identify video segments matching natural language descriptions; (b) \textit{Dense Video Captioning}~\cite{krishna2017dense, wang2021end}, which requires generating temporally aligned captions for multiple events; (c) \textit{Video Highlight Detection}~\cite{song2015tvsum, xiong2019less}, aims at selecting segments most relevant to a given query; and (d) \textit{Temporally Grounded Video Question Answering}~\cite{xiao2024can, chen2024rextime}, which involves pinpointing the temporal evidence needed to accurately answer questions. Collectively, these tasks define the contemporary scope of VTG research and highlight the necessity of sophisticated temporal reasoning. 

Although substantial progress has been achieved, early VTG methods based on traditional deep learning architectures~\cite{zhang2019man, moon2023query} continue to face significant limitations. These include challenges in bridging semantic gaps between visual and linguistic modalities, inadequate temporal context modeling, and limited generalization capabilities. Previous methods often relied on manually designed proposal-generation mechanisms~\cite{ma2020vlanet, Zhang_2021_CVPR} or simple temporal boundary regression~\cite{liu2022umt, yuan2019find}, which lacked flexibility and interpretability. Recently, the advent of Large Language Models (LLMs)~\cite{zhang2022opt, chung2024scaling, guo2025deepseek} and their multimodal variants, i.e., Multimodal Large Language Models (MLLMs)~\cite{liu2023visual, instructblip, damonlpsg2024videollama2, xu2024pllava}, has dramatically reshaped the field of video-language understanding. These models provide powerful cross-modal reasoning, instruction-following capabilities, and robust zero-shot generalization, significantly enhancing the potential for effective VTG.

\begin{figure*}[t]
\centering
  \includegraphics[width=\linewidth]{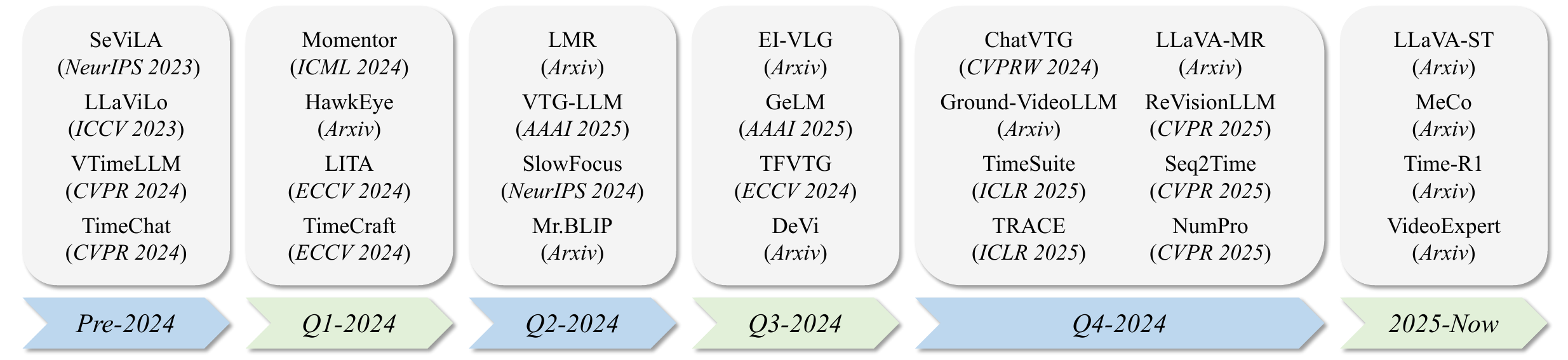}
\vspace{-0.2cm}
\caption{An overview timeline of representative VTG-MLLMs. This timeline is organized according to the initial arXiv release dates of each work, with corresponding conference acceptance details included where applicable.}
\label{Fig:timeline}
\vspace{-0.5cm}
\end{figure*}

Motivated by these advancements, a rapidly growing research direction termed \textit{VTG-MLLMs} \cite{huang2024vtimellm, ren2024timechat, qian2024momentor} has emerged, leveraging MLLMs for temporal grounding tasks. The swift evolution of this subfield is visually charted in Fig.~\ref{Fig:timeline}, which presents a chronological overview of representative VTG-MLLM approaches. Unlike traditional approaches relying solely on visual backbones with task-specific heads, VTG-MLLMs utilize general-purpose MLLMs to reason about temporal relationships, align semantics, and localize relevant video segments either directly or indirectly. VTG-MLLMs adopt diverse architectural strategies, with some methods employing MLLMs as high-level facilitators for semantic grounding \cite{qu2024chatvtg, qin2024question}, and others using them explicitly for boundary prediction~\cite{guo2024trace, wang2024hawkeye}. Consequently, the VTG-MLLM field now encompasses varied architectural innovations, training paradigms, and representation techniques~\cite{zeng2024timesuite, wang2024grounded, liu2025videomind}. However, the rapid evolution and complexity of VTG-MLLM research present challenges for navigating the current literature. Existing surveys predominantly focus on general video-language modeling~\cite{li2024multimodal, zhu2020comprehensive, abdar2024review, xing2024survey, zhang2024vision} or cover VTG from a pre-LLM perspective~\cite{liu2021survey, yang2020survey, lan2023survey, zhang2023temporal}, leaving a notable gap in systematic analyses of VTG in the LLM era.

% , covering literature up to April 2025
To fill this gap, we present the first comprehensive survey exclusively dedicated to VTG-MLLMs, covering literature up to May 2025. 
This survey systematically organizes recent developments, identifies emerging technical trends, and outlines future research opportunities. Specifically, we introduce a structured three-dimensional taxonomy:
\begin{itemize}
\item \textbf{Functional Roles of MLLMs}: Classifying models based on whether MLLMs act as \textit{Facilitators} assisting downstream grounding tasks, or as \textit{Executors} directly predicting temporal boundaries.
\item \textbf{Training Paradigms}: Distinguishing between \textit{pretraining}, \textit{fine-tuning}, and \textit{training-free} approaches, each with unique trade-offs in generalization, task specialization, and supervision requirements.
\item \textbf{Video Feature Processing Techniques}: Examining strategies for representing and integrating video inputs, including spatiotemporal tokenization and temporal modeling techniques.
\end{itemize}

Our taxonomy provides a progressive analytical framework, guiding readers from the high-level roles of MLLMs through training paradigms and down to video feature processing methods. By structuring our survey in this layered manner (see Fig.~\ref{Fig:taxo-model}), we aim to offer clarity, advance comparative analyses, and identify underexplored avenues in VTG-MLLMs.

The remainder of this survey is organized as follows. Section~\ref{preliminaries} introduces the preliminaries of VTG-MLLMs, including an overview of VTG tasks and the foundational concepts behind MLLMS. Section~\ref{Taxonomy} presents a detailed taxonomy of recent VTG-MLLM research, categorizing methods along three key dimensions: the \textit{functional roles} of MLLMs, \textit{training paradigms}, and \textit{video feature processing} strategies. These categories and their subtypes are illustrated in Fig.~\ref{Fig:taxo-model}. 
Section~\ref{benchmark_experiment} provides an overview of benchmark datasets, evaluation protocols, and comparative analysis of empirical results across existing VTG-MLLMs.
Section~\ref{future} discusses open challenges and future research directions. Finally, Section~\ref{conclusion} concludes the survey.

\section{Preliminaries}
\label{preliminaries}
In this section, we provide an overview of core VTG tasks, as well as the foundational background on MLLMs.

\subsection{Video Temporal Grounding}
\label{tasks}
In this survey, we provide a comprehensive overview of four primary VTG tasks: video moment retrieval, dense video captioning, video highlight detection, and temporally grounded video question answering. In the following subsections, we briefly describe each of these tasks.

\subsubsection{Video Moment Retrieval}\label{mr} 
Video moment retrieval (MR)~\cite{zhang2020learning, lin2023univtg}, also referred to as \textit{temporal sentence grounding}~\cite{yang2021local, liu2021context}, \textit{video moment localization}~\cite{xiao2021boundary, zhang2021natural} or \textit{temporal video grounding}~\cite{ren2024timechat, liu2024bench}, aims to identify and localize temporal segments within untrimmed videos based on natural language queries (see Fig.~\ref{Fig:Illustration-tasks}~(a)). This task represents the most direct and fundamental benchmark for evaluating the temporal grounding capabilities of VTG-MLLMs. It requires not only accurate alignment of textual descriptions with specific video segments, but also the ability to map video content to precise temporal boundaries, testing a model’s understanding of fine-grained temporal relationships and semantic coherence in untrimmed videos.

\subsubsection{Dense Video Captioning}\label{dc} 
\begin{figure*}[t]
\centering
\includegraphics[width=0.8\linewidth]{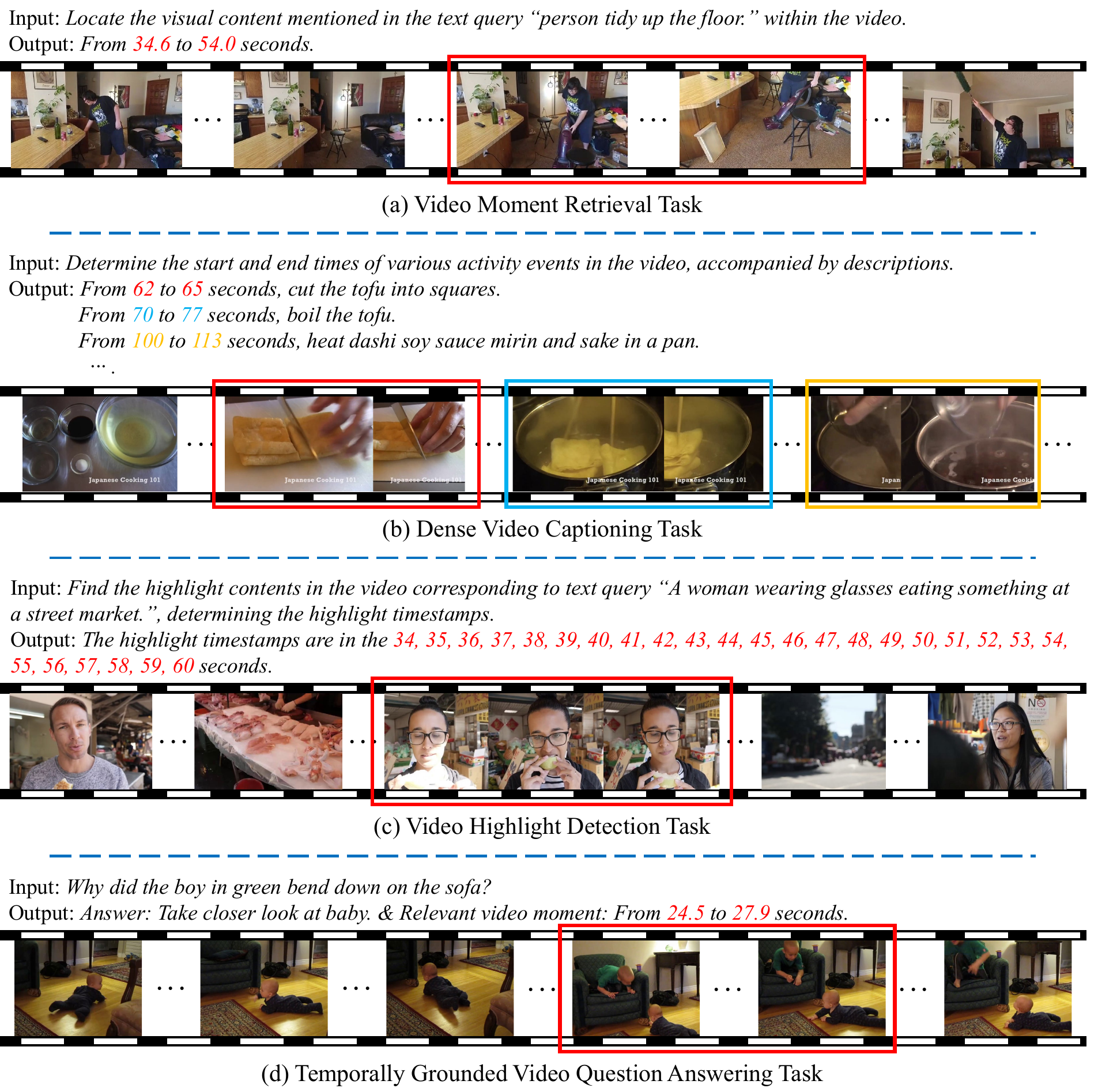}
\vspace{-0.2cm}
\caption{Illustration of four core tasks in VTG: video moment retrieval, dense video captioning, video highlight detection, and temporally grounded video question answering. These tasks encompass a range of temporal reasoning challenges, including precise segment localization, multi-event description generation, highlight identification, and time-sensitive question answering, each requiring fine-grained temporal understanding.}
\label{Fig:Illustration-tasks}
\vspace{-0.5cm}
\end{figure*}

Dense video captioning (DC)~\cite{zhou2018end, iashin2020multi} involves generating detailed, temporally grounded descriptions for all significant events or actions occurring in an untrimmed video, along with their corresponding start and end timestamps (see Fig.~\ref{Fig:Illustration-tasks}~(b)). Unlike MR, which aims at localizing a single specific moment given a textual query, DC captures the complete narrative by identifying multiple events and their intricate temporal dependencies. This task assesses a model's proficiency in comprehending extended temporal contexts and nuanced interactions within videos. Additionally, DC explicitly challenges models to manage overlapping events, a capability essential for achieving comprehensive fine-grained video understanding using large language models.

\subsubsection{Video Highlight Detection}\label{hd} 
Video highlight detection (HD)~\cite{lei2021detecting, song2015tvsum} aims to identify keyframes or short segments within an untrimmed video that best match a given textual query, typically by assigning an importance or relevance score to these moments (see Fig.~\ref{Fig:Illustration-tasks}~(c)). 
Unlike MR and DC, which primarily operate at the event level, HD emphasizes frame-level precision. This task evaluates the ability of model to accurately pinpoint salient video clips that closely correspond to textual prompts and to assess their contextual significance. Such fine-grained alignment is essential for applications that require high temporal precision in identifying critical events.

\subsubsection{Temporally Grounded Video Question Answering}\label{gqa} 
Temporally grounded video question answering (GQA)~\cite{chen2024cg, liu2024bench}, also known as \textit{grounded video QA}\cite{liu2025timecraft, xiao2024can} or \textit{temporal video grounding of questions}\cite{wang2024hawkeye}, extends traditional video QA by requiring models to not only answer questions but also identify and localize the precise temporal intervals containing relevant visual evidence (see Fig.~\ref{Fig:Illustration-tasks}~(d)). Unlike MR, GQA introduces the added complexity of integrating temporal localization with multimodal reasoning. This task is particularly critical for developing explainable video QA systems, as it demands explicit and interpretable connections between textual answers and visual evidence within the video content.

\subsection{Multimodal Large Language Models}

MLLMs \cite{zhang2023video, lin2023video, li2023videochat} extend traditional LLMs by integrating multimodal encoders, such as image encoders \cite{dosovitskiy2020image, sun2023eva, fang2023eva, radford2021learning}, video encoders \cite{feichtenhofer2022masked, tong2022videomae, wang2023videomae, wang2023all}, and specialized cross-modal adapters \cite{instructblip, sun2024video, azad2025hierarq}. Taking Video-LLM as an illustrative example, a video encoder processes a sequence of downsampled frames $V$, converting them into visual tokens $F^{v}=E_{v}(V)$. These visual tokens are then projected by a visual adapter to align with the embedding space of the language model, yielding aligned visual tokens $X^{v}=Q(F^{v})$. Concurrently, an input textual query $q$, which may include instructions, prompts, or other textual elements, is encoded into linguistic tokens $X^{t}$ via a textual encoder. The visual and textual tokens are concatenated into a unified input sequence $[X^{v}, X^{t}]$, which is subsequently processed by the LLM to generate the appropriate inference.

Current research efforts on MLLMs emphasize maximizing the efficiency of leveraging LLMs' advanced capabilities. Initial studies predominantly focus on designing cross-modality adapters aimed at mapping features from non-linguistic modalities into the semantic embedding space of language models. Flamingo \cite{alayrac2022flamingo}, as a pioneering model, integrates visual and linguistic modalities through a gated cross-attention mechanism. Subsequently, various approaches, such as BLIP \cite{li2023blip}, mPLUG \cite{ye2023mplug}, and LanguageBind \cite{huang2023language}, adopt the Q-Former architecture to align visual representations, whereas models in the LLaVA series \cite{liu2023visual, liu2024improved, li2024llava} introduce multilayer perceptrons (MLPs) as simpler yet effective connectors for modality integration. Additionally, more recent works have proposed lightweight and efficient alignment modules \cite{ye2024mplug3, chen2024evlm, shi2025slow}, continuing to enhance performance and model compactness.

Alongside architectural developments, training strategies constitute another critical research direction. Researchers have compiled large-scale multimodal pretraining datasets \cite{miech2019howto100m, bain2021frozen, schuhmann2022laion} to facilitate robust and diverse representation learning. Instruction-tuned datasets \cite{li2023m, yin2023lamm} and specialized datasets tailored for chain-of-thought (CoT) reasoning \cite{shao2024visual, gao2024cantor, wu2024visual} have been developed to improve task comprehension and generalization capabilities of MLLMs. Furthermore, parameter-efficient fine-tuning methods such as LoRA \cite{hu2022lora}, LISA \cite{pan2024lisa}, and DoRA \cite{liu2024dora} have emerged, enabling efficient task-specific or domain-specific adaptation without extensive retraining.

\tikzstyle{my-box}=[
    rectangle,
    draw=gray,
    rounded corners,
    text opacity=1,
    minimum height=2em,
    minimum width=5em,
    inner sep=2pt,
    align=center,
    fill opacity=.5,
    line width=0.8pt,
]
\tikzstyle{leaf}=[my-box, minimum height=2em,
    fill=hidden-pink!80, text=black, align=left,font=\normalsize,
    inner xsep=2pt,
    inner ysep=4pt,
    line width=0.8pt,
]

\definecolor{c1}{RGB}{179,217,255} % soft blue
\definecolor{c2}{RGB}{251,218,131} % soft orange
\definecolor{c3}{RGB}{189,233,189} % soft green
\definecolor{c4}{RGB}{153,221,214} % soft turquoise

\begin{figure*}[htp]
    \centering
    \resizebox{0.85\textwidth}{!}{
        \begin{forest}
            for tree={
                grow=east,
                reversed=true,
                anchor=base west,
                edge path={
                    \noexpand\path [\forestoption{edge}] (!u.parent anchor) -- +(5pt,0) |- (.child anchor)\forestoption{edge label};
                },
                parent anchor=east,
                child anchor=west,
                base=center,
                font=\large,
                rectangle,
                draw=hidden-draw,
                rounded corners,
                align=left,
                text centered,
                minimum width=6em,
                edge+={darkgray, line width=1pt},
                s sep=20pt,
                inner xsep=3pt,
                inner ysep=3pt,
                line width=0.8pt,
                ver/.style={rotate=90, child anchor=north, parent anchor=south, anchor=center, minimum width=10em},
            },
            where level=1{text width=9.7em,font=\normalsize,}{},
            where level=2{text width=8.2em,font=\normalsize,}{},
            where level=3{text width=4.0em,font=\normalsize,}{},
            where level=4{font=\normalsize,}{},
            [
                VTG-MLLMs, ver
                [
                    \shortstack{\textcolor{red}{$\S~$}Section~\ref{functional role} \\[0.4ex] Functional Roles \\ of MLLMs}, fill=c1!60, draw=c1, line width=0.5mm
                    [
                        Facilitator, fill=c1!60, draw=c1, edge={c1}, line width=0.5mm
                        [
                            ChatVTG \cite{qu2024chatvtg}{, }
                            TEA \cite{cai2024mllm}{, }
                            TimeCraft \cite{liu2025timecraft}{, }
                            GroundVQA \cite{di2024grounded}{, }\\
                            TFVTG \cite{zheng2025training}{, }
                            Grounding-Prompter \cite{chen2023grounding}{, }
                            VTG-GPT \cite{xu2024vtg}{, }\\
                            VERIFIED \cite{chen2024verified}{, }
                            EI-VLG \cite{lee2024infusing}{, }
                            VideoLights \cite{paul2024videolights}{, }
                            GPTSee \cite{sun2023gptsee}{, }\\
                            LMR \cite{liu2024context}{, }
                            ReCorrect \cite{bao2024vid}{, }
                            DeVi \cite{qin2024question}{, }
                            Moment-GPT \cite{xu2025zero}{ }
                            , leaf, text width=31.1em, draw=c1, edge={c1}, line width=0.7mm
                        ]
                    ]
                    [
                        Executor, fill=c1!60, draw=c1, edge={c1}, line width=0.5mm
                        [
                            SeViLA \cite{yu2023self}{, }
                            LLaViLo \cite{ma2023llavilo}{, }
                            TimeChat \cite{ren2024timechat}{, }
                            MLLM-TA \cite{liu2024mllm}{, }\\
                            LLaVA-MR \cite{lu2024llava}{, }
                            NumPro \cite{wu2024number}{, }
                            TRACE \cite{guo2024trace}{, }
                            VTimeLLM \cite{huang2024vtimellm}{, }\\
                            Seq2Time \cite{deng2024seq2time}{, }
                            Grounded-VideoLLM \cite{wang2024grounded}{, }
                            TimeSuite \cite{zeng2024timesuite}{, }\\
                            VTG-LLM \cite{guo2024vtg}{, }
                            GroundingGPT \cite{li2024groundinggpt}{, }
                            ReVisionLLM \cite{hannan2024revisionllm}{, }\\
                            LITA \cite{huang2025lita}{, }
                            Momentor \cite{qian2024momentor}{, }
                            HawkEye \cite{wang2024hawkeye}{, }
                            TimeRefine \cite{wang2024timerefine}{, }\\
                            LLaVA-ST \cite{li2025llava}{, }
                            TimeMarker \cite{chen2024timemarker}{, }
                            GeLM \cite{chen2024grounded}{, }
                            SlowFocus \cite{nie2024slowfocus}{, }\\
                            Mr.BLIP \cite{meinardus2024surprising}{, }
                            TemporalVLM \cite{fateh2024video}{, }
                            TGB \cite{wang2024efficient}{, }
                            TPE-VLLM \cite{li2025mitigating}{, }\\
                            VideoChat-TPO \cite{yan2024task}{, }
                            SpaceVLLM \cite{wang2025spacevllm}{, }
                            E.T.Chat \cite{liu2024bench}{, }\\
                            MeCo \cite{pang2025measure}{, }
                            Time-R1 \cite{wang2025timezero}{, }
                            VideoExpert \cite{zhao2025videoexpert}{, }
                            VideoMind \cite{liu2025videomind}{, }\\
                            VideoChat-R1 \cite{li2025videochat}{, }
                            MUSEG \cite{luo2025museg}{ }
                            , leaf, text width=31.1em, draw=c1, edge={c1}, line width=0.7mm
                        ]
                    ]
                ]
                [                
                    \shortstack{\textcolor{red}{$\S~$}Section~\ref{training paradigms} \\[0.4ex] Training Paradigms}, fill=c2!60, draw=c2, line width=0.5mm
                    [
                        Pretraining, fill=c2!60, draw=c2, edge={c2}, line width=0.5mm
                        [
                            SeViLA \cite{yu2023self}{, }
                            VTimeLLM \cite{huang2024vtimellm}{, }
                            TimeChat \cite{ren2024timechat}{, }
                            VTG-LLM \cite{guo2024vtg}{, }\\
                            LITA \cite{huang2025lita}{, }
                            TemporalVLM \cite{fateh2024video}{, }
                            Seq2Time \cite{deng2024seq2time}{, }
                            TimeSuite \cite{zeng2024timesuite}{, }\\
                            VideoChat-TPO \cite{yan2024task}{, }
                            Grounded-VideoLLM \cite{wang2024grounded}{, }
                            HawkEye \cite{wang2024hawkeye}{, }\\
                            TRACE \cite{guo2024trace}{, }
                            GroundingGPT \cite{li2024groundinggpt}{, }
                            MLLM-TA \cite{liu2024mllm}{, }
                            GeLM \cite{chen2024grounded}{, }\\
                            TimeRefine \cite{wang2024timerefine}{, }
                            TimeMarker \cite{chen2024timemarker}{, }
                            Momentor \cite{qian2024momentor}{, }\\
                            ReVisionLLM \cite{hannan2024revisionllm}{, }
                            SlowFocus \cite{nie2024slowfocus}{, }
                            LLaVA-ST \cite{li2025llava}{, }\\
                            TPE-VLLM \cite{li2025mitigating}{, }
                            E.T.Chat \cite{liu2024bench}{, }
                            MeCo \cite{pang2025measure}{, }
                            VideoMind \cite{liu2025videomind}{, }\\
                            SpaceVLLM \cite{wang2025spacevllm}{, }
                            VideoExpert \cite{zhao2025videoexpert}{, }
                            Time-R1 \cite{wang2025timezero}{, }\\
                            VideoChat-R1 \cite{li2025videochat}{, }
                            MUSEG \cite{luo2025museg}{ }
                            , leaf, text width=31.1em, draw=c2, edge={c2}, line width=0.7mm
                        ]
                    ]
                    [
                        Fine-Tuning, fill=c2!60, draw=c2, edge={c2}, line width=0.5mm
                        [
                            LLaViLo \cite{ma2023llavilo}{, }
                            Mr.BLIP \cite{meinardus2024surprising}{, }
                            LLaVA-MR \cite{lu2024llava}{, }
                            VideoLights \cite{paul2024videolights}\\
                            TGB \cite{wang2024efficient}{, }
                            GPTSee \cite{sun2023gptsee}{, }
                            EI-VLG \cite{lee2024infusing}{, }
                            LMR \cite{liu2024context}{, }
                            TEA \cite{cai2024mllm}{, }\\
                            ReCorrect \cite{bao2024vid}{ }
                            , leaf, text width=31.1em, draw=c2, edge={c2}, line width=0.7mm
                        ]
                    ]
                    [
                        Training-Free, fill=c2!60, draw=c2, edge={c2}, line width=0.5mm
                        [
                            NumPro \cite{wu2024number}{, }
                            ChatVTG \cite{qu2024chatvtg}{, }
                            DeVi \cite{qin2024question}{, }
                            TFVTG \cite{zheng2025training}{, }\\
                            VTG-GPT \cite{xu2024vtg}{, }
                            Moment-GPT \cite{xu2025zero}{, }
                            Grounding-Prompter \cite{chen2023grounding}{ }
                            , leaf, text width=31.1em, draw=c2, edge={c2}, line width=0.7mm
                        ]
                    ]
                ]
                [
                    \shortstack{\textcolor{red}{$\S~$}Section~\ref{video feature extraction} \\[0.4ex] Feature Processing \\ Techniques}, fill=c3!60, draw=c3, line width=0.5mm
                    [
                        Visual Feature, fill=c4!60, draw=c4, edge={c4}, line width=0.5mm
                        [
                            With Compression, text width=8.5em, fill=c4!60, draw=c4, edge={c4}, line width=0.5mm
                            [
                                ReVisionLLM \cite{hannan2024revisionllm}{, }
                                LLaVA-MR \cite{lu2024llava}{, }\\
                                E.T.Chat \cite{liu2024bench}{, }
                                VTG-LLM \cite{guo2024vtg}{, }
                                LITA \cite{huang2025lita}{, }\\
                                Grounded-VideoLLM \cite{wang2024grounded}{, }
                                TRACE \cite{guo2024trace}{, }\\
                                TimeMarker \cite{chen2024timemarker}{, }
                                LLaVA-ST \cite{li2025llava}{, }\\
                                TimeSuite \cite{zeng2024timesuite}{, }
                                VideoExpert \cite{zhao2025videoexpert}{ }
                                , leaf, text width=20.7em, draw=c4, edge={c4}, line width=0.7mm
                            ]
                        ]
                        [
                            With Refinement, text width=8.5em, fill=c4!60, draw=c4, edge={c4}, line width=0.5mm
                            [
                                SeViLA \cite{yu2023self}{, }
                                HawkEye \cite{wang2024hawkeye}{, }
                                VideoMind \cite{liu2025videomind}{, }\\
                                SlowFocus \cite{nie2024slowfocus}{, }
                                ReVisionLLM \cite{hannan2024revisionllm}{ }
                                , leaf, text width=20.7em, draw=c4, edge={c4}, line width=0.7mm
                            ]
                        ]
                    ]
                    [
                        Temporal  Feature, fill=c3!60, draw=c3, edge={c3}, line width=0.5mm
                        [
                            Explicit, text width=4.9em, fill=c3!60, draw=c3, edge={c3}, line width=0.5mm
                            [
                                LLaVA-MR \cite{lu2024llava}{, }
                                TRACE \cite{guo2024trace}{, }
                                LITA \cite{huang2025lita}{, }\\
                                VTG-LLM \cite{guo2024vtg}{, }
                                Mr.BLIP \cite{meinardus2024surprising}{, }
                                Momentor \cite{qian2024momentor}{, }\\
                                TimeMarker \cite{chen2024timemarker}{, }
                                SlowFocus \cite{nie2024slowfocus}{, }\\
                                LLaVA-ST \cite{li2025llava}{, }
                                TGB \cite{wang2024efficient}{, }
                                VideoExpert \cite{zhao2025videoexpert}{ }
                                , leaf, text width=23.82em, draw=c3, edge={c3}, line width=0.7mm, draw=c3, edge={c3}, line width=0.7mm
                                % +2 ori 19.72 % +2
                            ]
                        ]
                        [
                            Implicit, text width=4.9em, fill=c3!60, draw=c3, edge={c3}, line width=0.5mm
                            [   
                                % Intrinsic \\ Reasoning, text width=4.5em, align=left
                                % [
                                    NumPro \cite{wu2024number}{, }
                                    HawkEye \cite{wang2024hawkeye}{, }
                                    ReVisionLLM \cite{hannan2024revisionllm}{, }\\
                                    GroundingGPT \cite{li2024groundinggpt}{, }
                                    GeLM \cite{chen2024grounded}{, }
                                    LLaViLo \cite{ma2023llavilo}{, }\\
                                    TimeRefine \cite{wang2024timerefine}{, }
                                    VTimeLLM \cite{huang2024vtimellm}{, }
                                    TimeChat \cite{ren2024timechat}{, }\\
                                    Grounded-VideoLLM \cite{wang2024grounded}{, }
                                    % , leaf, text width=16.45em
                            %     ]
                            % ]
                            % [   
                            %     Feature \\ Infusion, text width=4.5em, align=left
                            %     [
                                    TemporalVLM \cite{fateh2024video}{, }\\
                                    TimeSuite \cite{zeng2024timesuite}{, }
                                    TPE-VLLM \cite{li2025mitigating}{, }
                                    VideoMind \cite{liu2025videomind}{ }
                                    % , leaf, text width=16.45em
                                    , leaf, text width=23.82em, draw=c3, edge={c3}, line width=0.7mm, draw=c3, edge={c3}, line width=0.7mm
                                ]
                            ]
                        ]
                    ]
                ]
            ]
        \end{forest}
    }
    \caption{Taxonomy of VTG-MLLMs, encompassing three primary dimensions: \textit{Functional Roles of MLLMs} (Facilitators / Executors), \textit{Training Paradigms} (Pretraining / Fine-Tuning / Training-Free), and \textit{Feature Processing Techniques} (Visual Features / Temporal Features), each with distinct sub-categories reflecting the diverse strategies employed in this field.}
    \label{Fig:taxo-model}
\end{figure*}
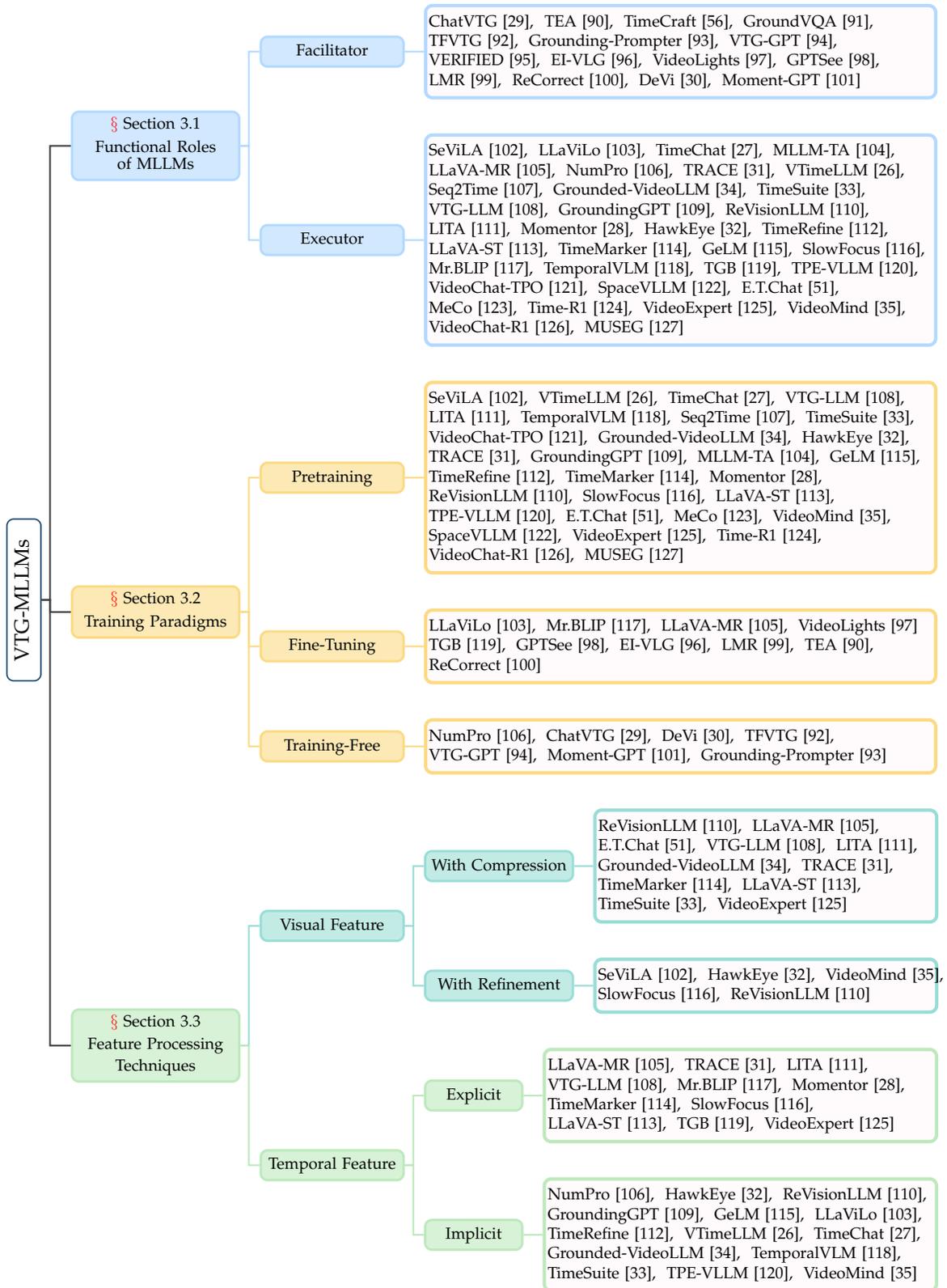

% \section{A Multi-faceted Taxonomy of VTG-MLLMs}
\section{A\,Multi-faceted\,Taxonomy\,of\,VTG-MLLMs}
\label{Taxonomy}

As established in the Introduction (Section~\ref{sec:introduction}), we employ a three-dimensional taxonomy to deconstruct the complexities of VTG-MLLMs. This section delves into the specifics of this classification. 
Our taxonomy (visualized in Fig.~\ref{Fig:taxo-model}), which progresses from high-level architectural considerations to fine-grained processing techniques, will be explored in detail through the following dimensions:
\begin{itemize}
\item \textbf{The Functional Roles of MLLMs} (Section~\ref{functional role}): We will analyze how the architectural positioning of MLLMs—whether as \textit{Facilitators} aiding downstream tasks or as \textit{Executors} directly undertaking temporal prediction—shapes their overall design and impact on temporal perception.
\item \textbf{The Training Paradigms} (Section~\ref{training paradigms}): This subsection will differentiate among the \textit{pretraining}, \textit{fine-tuning}, and \textit{training-free} paradigms. 
The analysis will center on the inherent trade-offs each strategy presents in terms of generalization capability, task-specific adaptation, and overall resource demands.
\item \textbf{The Video Feature Processing Techniques} (Section~\ref{video feature extraction}): Here, we will systematically examine the diverse methodologies for representing and integrating video inputs. This includes a closer look at spatiotemporal tokenization mechanisms within the token budget, and various temporal modeling approaches that enable models to effectively process and reason about dynamic visual content.
\end{itemize}
This structured examination will provide the foundation for our detailed review of specific methodologies and trends within the VTG-MLLM field in the sections that follow.

\subsection{Functional Roles of MLLMs in VTG-MLLMs}
\label{functional role}
The functional role of MLLMs characterizes their architectural integration within VTG pipelines, determining whether they function primarily as auxiliary modules facilitating cross-modal understanding or as central reasoning engines directly conducting temporal grounding. Accordingly, existing VTG-MLLMs can be categorized into two paradigms: 1) \textit{Facilitators}, where MLLMs generate structured textual representations from video content to support downstream modules; and 2) \textit{Executors}, where MLLMs directly perform temporal boundary prediction via integrated multimodal reasoning.

\begin{figure*}[t]
\centering
   \includegraphics[width=0.85\linewidth]{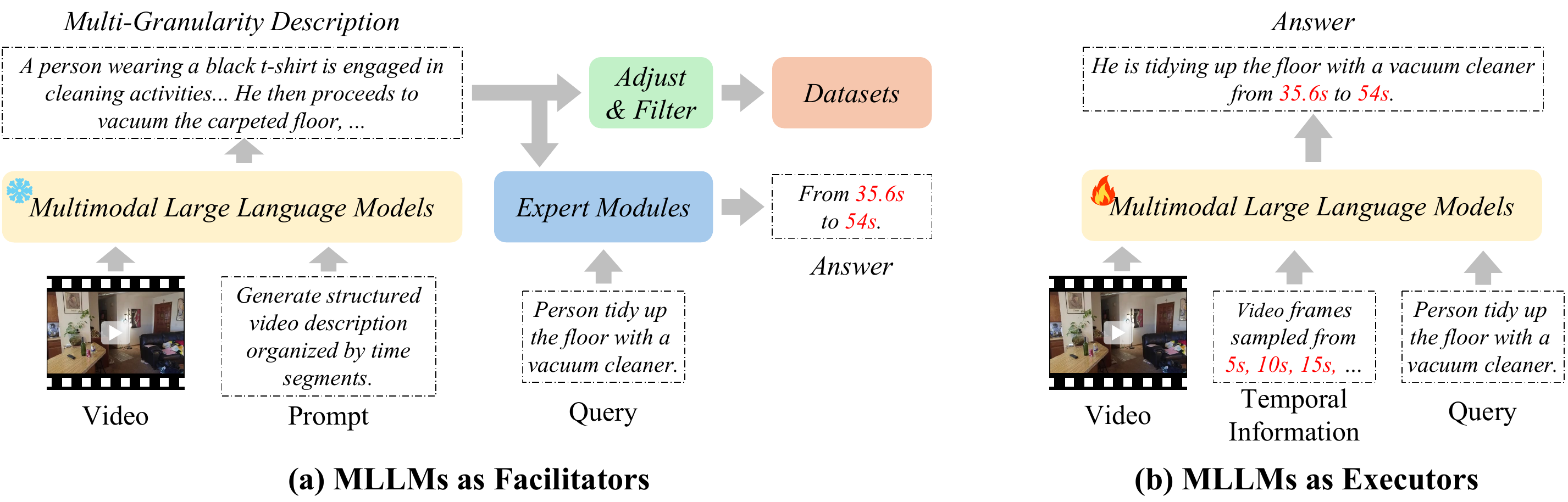}
\vspace{-0.2cm}
\caption{Visualization of the distinct functional roles of MLLMs in VTG-MLLMs. The left panel depicts MLLMs as \textit{Facilitators}, where they generate textual descriptions from video inputs, serving either as training data for downstream tasks or as auxiliary signals within expert modules. The right panel illustrates MLLMs as \textit{Executors}, directly processing queries, video content, and temporal information (e.g., timestamps) to produce grounded outputs through a sequence-to-sequence prediction framework.}
\label{Fig:mllms' role}
\vspace{-0.5cm}
\end{figure*}

% \subsection{MLLM as Facilitator}
\subsubsection{Facilitators}

In the Facilitator role, MLLMs act as intermediaries by transforming complex video data into structured textual forms, as depicted in Fig.~\ref{Fig:mllms' role}~(a). 
We formalize this process as a conditional generation problem: 
\begin{equation}
T = \text{MLLM}_\text{facilitator}(V),
\end{equation}
where $V$ denotes the input video and $T$ represents the generated textual descriptions. 
The generated textual outputs can either directly facilitate dataset construction or serve as semantic aids within dedicated downstream modules. Two primary application areas arise under this paradigm: \textit{Dataset Construction} and \textit{Expert Module Integration}.

\textbf{Dataset Construction}: MLLMs are extensively utilized to synthesize textual annotations, significantly enhancing the efficiency of dataset creation and expansion for model training and evaluation. For instance, \citet{di2024grounded} leverages Llama2~\cite{touvron2023llama} to transform timestamped narrations from Ego4D~\cite{grauman2022ego4d} into temporally grounded QA pairs. Similarly, GPT-4o~\cite{openai2024gpt4o} was employed by \citet{bao2024vid} in Vid-Morp to automatically generate pseudo-labeled sentences aligned with video frames. Other works~\cite{paul2024videolights, chen2024verified} have similarly utilized advanced models like BLIP-2~\cite{li2023blip}, LLaVA~\cite{liu2023visual}, and Gemini-1.5~\cite{team2024gemini} to automate the annotation process and enrich datasets for VTG tasks.

\textbf{Expert Module Integration}: Beyond dataset generation, textual outputs from MLLMs can serve directly within VTG systems, either as semantic inputs in similarity-based grounding methods or as additional signals enhancing visual representations through cross-modal integration. For instance, \citet{qu2024chatvtg} employs Video-ChatGPT~\cite{maaz2023video} to generate multi-granularity clip captions, facilitating iterative query matching using Sentence-BERT~\cite{reimers2019sentence}. Similarly, \citet{xu2024vtg} utilizes MiniGPT-v2~\cite{chen2023minigpt} for caption generation and Baichuan2~\cite{yang2023baichuan} for query rewriting, reducing linguistic biases. Additionally, \citet{cai2024mllm} leverages LLaVA-1.5~\cite{liu2024improved} to generate paragraph-level narrations, aligning them temporally with video features via cross-attention mechanisms~\cite{moon2023query, liu2022umt}, thereby enhancing contextual understanding and robustness.

\textbf{Summary}: The Facilitator framework is advantageous due to its computational efficiency, ease of deployment, and inherent scalability, requiring minimal adaptation of pretrained MLLMs. However, reliance on static pretrained models carries limitations such as propagation of inherent biases~\cite{fraser2024examining} and constraints from original training data, potentially affecting the reliability of textual outputs and downstream performance. Furthermore, the fixed nature of off-the-shelf MLLMs inherently limits their capability for complex temporal reasoning, creating performance bottlenecks that sophisticated pipeline designs may not fully address. Nonetheless, as Facilitators, MLLMs remain valuable for efficient dataset curation, providing abundant task-specific data crucial for advancing VTG research.

\subsubsection{Executors}
\label{role: executor}
When functioning as Executors, MLLMs directly perform the core tasks of VTG, formulating the problem as an end-to-end sequence-to-sequence (seq-to-seq) prediction challenge. 
In this setting, illustrated in Fig.~\ref{Fig:mllms' role}~(b), the model jointly consumes raw video input and task-specific textual prompts to generate a temporally aligned output:
\begin{equation}
Y = \text{MLLM}_\text{executor}(V, Q, \tau),
\end{equation}
where $V$ denotes the input video stream, $Q$ is the textual query, and $\tau$ optionally encapsulates temporal priors. The output $Y$ represents predicted answers, timestamps, or task-specific tokens aligned with the video timeline.  

This paradigm holds the potential to unify video understanding within a generative framework. However, it faces a significant hurdle: standard MLLMs~\cite{maaz2023video, li2023videochat} often struggle to capture fine-grained temporal dependencies. This challenge largely stems from their vision encoders treating video as an unordered "bag-of-features," which discards the crucial sequential information essential for precise event localization. To address these challenges, research has focused on two primary directions: \textit{Architectural Enhancement} and \textit{Training Optimization}.

\textbf{Architectural Enhancement}: Architectural innovations are designed to improve the temporal perception and reasoning capabilities of MLLMs. These enhancements typically involve either modifying the input feature processing pipeline or integrating temporal awareness directly into the LLM's internal structure.

One line of research focuses on enhancing video feature representations before they are processed by the LLM. For instance, the method in Momentor~\cite{qian2024momentor} injects explicit temporal position encodings into frame-level features to improve temporal localization. Grounded-VideoLLM~\cite{wang2024grounded} adopts a dual-stream architecture to separately capture spatial and temporal dynamics, while LLaVA-MR~\cite{lu2024llava} introduces components for reducing redundancy and emphasizing critical dynamic moments. These strategies aim to provide the LLM with richer temporal context, laying a stronger foundation for subsequent reasoning.

A complementary approach modifies the internal architecture or output mechanisms of LLMs to better handle temporal cues. For example, GeLM~\cite{chen2024grounded} incorporates flexible grounding tokens for temporal evidence retrieval, and TRACE~\cite{guo2024trace} adds task-specific decoding heads for structured temporal output. VideoExpert~\cite{zhao2025videoexpert} integrates parallel reasoning and generation modules for specialized processing. Similarly, VideoMind~\cite{liu2025videomind} decomposes complex tasks into specialized roles, like a Grounder with a timestamp-decoder and a Planner to coordinate them, using a Chain-of-LoRA strategy for seamless collaboration. While these approaches can significantly enhance temporal understanding, they may introduce trade-offs, such as increased computational overhead or reduced general-purpose flexibility.

\textbf{Training Optimization}: 
Effective optimization strategies are critical for equipping MLLMs with robust temporal understanding, transforming them into capable Executors.
These strategies typically form a holistic pipeline combining novel training curricula with temporal tasks and datasets.

A prevalent approach involves multi-stage training frameworks, as seen in VTimeLLM~\cite{huang2024vtimellm} and SlowFocus~\cite{nie2024slowfocus}, which progressively refine the model's temporal localization abilities. Another emerging direction is optimization via reinforcement learning (RL). Time-R1~\cite{wang2025timezero} adapts RL-based strategies specifically for temporal reasoning, and VideoChat-R1~\cite{li2025videochat} explores the effectiveness of Group Relative Policy Optimization (GRPO)~\cite{shao2024deepseekmath}.

To further strengthen temporal perception, many methods incorporate explicit reasoning tasks. For example, TPE-VLLM~\cite{li2025mitigating} introduces novel training objectives targeting boundary detection and duration reasoning, improving its handling of complex temporal relationships.

Underpinning all these strategies is the reliance on high-quality, time-annotated datasets. For instance, the TimeIT dataset from TimeChat~\cite{ren2024timechat} provides rich timestamp annotations essential for instruction tuning, while TimeSuite~\cite{zeng2024timesuite} offers a unified collection of diverse datasets to facilitate more comprehensive temporal learning.

\textbf{Summary}: The Executor paradigm represents a pivotal shift towards unified, end-to-end temporal grounding, allowing MLLMs to simultaneously process video content and textual queries within a tightly coupled seq-to-seq framework. This approach supports flexible input and output formats, capturing complex visual-textual correlations without relying on modular, cascaded architectures. However, this flexibility comes at a cost, often requiring extensive annotated datasets, significant computational resources, and complex training procedures. Despite these challenges, the Executor approach remains a promising direction for advancing fine-grained video understanding, with the potential to fundamentally reshape the field by integrating deeper temporal reasoning into multimodal models.

\subsection{Training Paradigms of VTG-MLLMs}
\label{training paradigms}
Building on the functional differentiation of MLLMs, this subsection examines the training paradigms used to adapt these models for effective video temporal grounding. The choice of training approach reflects not only the system's design goals, i.e., whether to build a domain-generalist or task-specific model, but also the trade-offs in supervision, resource efficiency, and scalability. We categorize current VTG-MLLM approaches into three main paradigms: \textit{Pretraining}, \textit{Fine-Tuning}, and \textit{Training-Free} pipelines.

\subsubsection{Pretraining VTG-MLLMs}
\label{pretraining subsec}
Pretraining in VTG-MLLMs aims to equip models with robust temporal reasoning capabilities via large-scale supervised learning. 
Fundamentally, like most generative multimodal approaches, the pretraining process involves training the model to generate a target output $T$, typically encompassing temporal annotations such as event order, timestamps, or durations, conditioned on an input video  $V$ and textual prompt $P$. Formally, this objective is represented as minimizing the pretraining loss over a dataset $\mathcal{D}$:
\begin{equation} \label{eq:unified_gen_loss}
\mathcal{L}_{\text{pretrain}} = \sum_{(V, T) \in \mathcal{D}} \mathcal{L}_{\text{gen}}(T \mid V, P).
\end{equation}
However, unlike general-purpose MLLMs, the primary innovation in pretraining-based VTG-MLLMs lies not in architectural design but in the creation of sophisticated training strategies and specialized pretraining datasets that tailor the optimization of Eqn.~(\ref{eq:unified_gen_loss}) for complex temporal understanding. We focus on two critical aspects of this approach: \textit{Prevalent Pretraining Strategies} and \textit{High-Quality Temporal Datasets}.

\textbf{Prevalent Pretraining Strategies}: 
The cornerstone of VTG pretraining is the multi-stage, progressive supervised learning pipeline. As introduced in the context of Executor models (Section~\ref{role: executor}), this strategy is founded on the principle of incremental learning, guiding the model from coarse-grained understanding to fine-grained localization. For instance, VTimeLLM~\cite{huang2024vtimellm} exemplifies this with its boundary-aware three-stage process, which sequentially tackles feature alignment, instruction tuning, and precise boundary optimization. Similarly, SlowFocus~\cite{nie2024slowfocus} integrates mixed-frequency sampling in its final training stages to enhance temporal resolution. This multi-stage philosophy has become a de facto standard, with numerous other models like TimeMarker~\cite{chen2024timemarker}, GroundingGPT~\cite{li2024groundinggpt}, and LLaVA-ST~\cite{li2025llava} adopting similar hierarchical frameworks to progressively refine temporal perception.

While multi-stage supervision remains the mainstream, a new wave of research is exploring innovative refinements, aiming for greater precision and efficiency. A particularly prominent direction is the application of RL. Inspired by the success of RL and techniques like GRPO~\cite{shao2024deepseekmath} in complex reasoning domains such as code generation and mathematics, researchers have begun adapting these methods for VTG. RL allows for the direct optimization of task-specific metrics like IoU, achieved by designing a composite reward function that encourages both a structured reasoning process and high prediction accuracy.
Pioneering this direction, Time-R1~\cite{wang2025timezero} introduces a reasoning-guided framework with a novel reward mechanism; VideoChat-R1~\cite{li2025videochat} further offers a systematic exploration with GRPO, while MUSEG~\cite{luo2025museg} addresses the single-segment limitation by enabling reasoning over multiple distributed events.

Beyond these, other novel strategies are also emerging. For example, Seq2Time~\cite{deng2024seq2time} adopts a data-centric strategy, synthesizing sequential training data with self-generated temporal cues, while TimeRefine~\cite{wang2024timerefine} reformulates temporal grounding as an iterative refinement task, allowing model to self-improve its localization accuracy.

\textbf{High-Quality Temporal Datasets}: 
High-quality, temporally annotated multimodal datasets are crucial for pretraining and instruction-tuning VTG models, providing the diverse contexts required for robust generalization. Building upon early efforts like TimeIT~\cite{ren2024timechat} and VTimeLLM~\cite{huang2024vtimellm}, subsequent work has evolved along several key directions. 
One major line of work aims to enhance data scale and diversity for more effective pretraining. For instance, VTG-IT-120K~\cite{guo2024vtg} expands on TimeIT by incorporating annotations from YT-Temporal-180M~\cite{zellers2021merlot}, InternVid-G~\cite{wang2024hawkeye} enriches InternVid10M-FLT~\cite{wang2023internvid} with segment-level captions and hard negative samples for more precise grounding, and Vid-Morp~\cite{bao2024vid} leverages pseudo-labeling on real-world videos to scale data creation. 
A parallel direction develops specialized instruction-tuning datasets to align models with complex temporal reasoning. Moment-10M~\cite{qian2024momentor}, sampled from YT-Temporal-1B~\cite{zellers2022merlot}, was designed for this purpose, while E.T. Instruct 164K~\cite{liu2024bench} complements this by providing a curated dataset across nine distinct tasks, specifically tailored for multi-event and time-sensitive scenarios. 
More recently, frontier datasets have begun to broaden the scope of VTG by integrating spatial dimensions~\cite{wang2025spacevllm} or advancing spatiotemporal understanding~\cite{li2025llava}.
Collectively, these datasets, along with numerous other contributions~\cite{zeng2024timesuite, wang2024grounded, wang2025timezero}, underscore a clear trajectory toward building more comprehensive and fine-grained data resources to tackle the full spectrum of temporal reasoning challenges.

\textbf{Summary}: The pretraining paradigm empowers MLLMs with robust temporal grounding capabilities, supporting generalization across diverse downstream VTG tasks. However, this approach also presents significant challenges, including the high computational cost of training and the substantial effort required to construct large, high-quality temporal datasets. Effective pretraining strategies must carefully balance task complexity and learning progression to maximize temporal understanding, making this a critical area for ongoing research.

\subsubsection{Fine-Tuning VTG-MLLMs}
\label{fine-tuning subsection}
In contrast to the resource-intensive pretraining paradigm (Section~\ref{pretraining subsec}), fine-tuning VTG-MLLMs provides a more computationally efficient approach, requiring smaller, task-specific datasets. Research within this paradigm can be broadly divided into two main directions: \textit{Direct Fine-Tuning of MLLMs} and \textit{Offline Textualization with MLLMs}, closely aligning with the functional roles discussed in Section~\ref{functional role}.

\textbf{Direct Fine-Tuning of MLLMs}: This approach directly fine-tunes general-purpose pre-trained MLLMs while maintaining their original architecture, adapting them for VTG tasks through task-specific training objectives. These methods typically reframe VTG as a seq-to-seq prediction problem, leveraging the contextual understanding capabilities already embedded in the models. 

For instance, SeViLA~\cite{yu2023self} adapts BLIP-2~\cite{li2023blip} into two interconnected components—a localizer and an answerer—strategically linking the outputs of the localizer to guide the answerer, thereby enhancing temporal precision. Similarly, LLaViLo~\cite{ma2023llavilo} incorporates lightweight adapters to integrate video-text features, utilizing a multi-objective loss function for more refined temporal grounding. TGB~\cite{wang2024efficient} introduces a unique approach by leveraging CNN-extracted optical flow features as low-dimensional motion cues, enhancing temporal awareness without substantially increasing model complexity. Additionally, recent innovations have focused on optimizing BLIP-2 for fine-grained temporal understanding. Notably, Mr.BLIP~\cite{meinardus2024surprising} explores novel multimodal input sequences to improve temporal understanding of events, while LLaVA-MR~\cite{lu2024llava} introduces a dynamic token compression strategy, reducing redundancy in spatiotemporal features and capturing more fine-grained event cues.

Despite their efficiency, these methods face the  \textit{catastrophic forgetting} challenge~\cite{zhai2023investigating}. As the models adapt to fine-tuning data, they often lose the general-purpose capabilities acquired during pretraining, leading to performance degradation on broader video understanding tasks.

\textbf{Offline Textualization with MLLMs}: An alternative fine-tuning strategy employs MLLMs in a static capacity to convert raw video inputs into textual descriptions, which then guide downstream modules. This approach effectively bridges the gap between unstructured visual data and language-conditioned learning tasks, often integrating components from traditional VTG methods.

For example, GPTSee~\cite{sun2023gptsee} generates detailed video descriptions, subsequently matched with textual queries to support moment localization. EI-VLG~\cite{lee2024infusing} incorporates these descriptions as environmental cues in a contrastive learning framework, refining the temporal precision of candidate segments. To address redundancy, LMR~\cite{liu2024context} uses cross-attention to highlight query-relevant segments, improving contextual alignment. Similarly, TEA~\cite{cai2024mllm} integrates these textual outputs with visual features to enhance semantic discriminability and temporal precision.

While this approach can significantly improve temporal grounding accuracy, it inherits certain limitations from traditional VTG methods. For instance, early two-stage matching approaches~\cite{chen2020learning, wang2021structured} often rely on pre-defined segment boundaries, limiting global context modeling. Meanwhile, direct regression methods~\cite{wang2020temporally, chen2021end} can misinterpret visually similar segments as semantically identical due to attention biases. In contrast, fine-tuned MLLMs provide richer, more context-aware embeddings, aligning visually similar events with distinct textual semantics and improving overall robustness.

\textbf{Summary}: Fine-tuning VTG-MLLMs offers a practical compromise between pretraining and fully training-free approaches, significantly reducing computational overhead while enhancing task-specific temporal alignment. However, these methods are inherently task-optimized, limiting their generalization across broader video understanding domains. As a result, fine-tuned models often excel in narrow, well-defined tasks but struggle with broader generalization—a critical challenge for future research.

\subsubsection{Training-Free VTG-MLLMs} 
\label{training-free subsec}
Training-free approaches represent a rapidly emerging paradigm in VTG-MLLMs, notable for their low computational overhead and zero-shot nature that eliminates the need for labeled supervision. These methods bypass the need for end-to-end training by leveraging pre-trained foundation models (e.g., MLLMs and LLMs) and specialized expert tools~\cite{li2023blip, reimers2019sentence, chen2022beats}, enabling temporal grounding through purely inference-based pipelines. While they share architectural similarities with fine-tuning approaches (Section~\ref{fine-tuning subsection}), training-free methods distinguish themselves by substituting trainable components with off-the-shelf models, significantly reducing the need for task-specific parameter updates. Current training-free VTG-MLLMs generally adopt one of two principal strategies: \textit{Feature Similarity Matching} and \textit{LLM-Driven Reasoning}, distinguished by their approach to leveraging MLLM-generated textualizations for temporal localization.

\textbf{Feature Similarity Matching}: This strategy relies on extracting high-dimensional semantic representations from both natural language queries and textualized video content using pre-trained encoders. 
Temporal grounding is then achieved by identifying the video span $s^*$ that maximizes a similarity score with the query $\mathbf{q}$, formalized as:
\begin{equation}
\label{eq:feature_similarity}
s^* = \underset{s_i \in \mathcal{S}}{\arg\max} \ \text{sim}(E_Q(\mathbf{q}), E_V(\mathbf{v_{s_i}})),
\end{equation}
where $\mathcal{S} = \{s_1, s_2, \dots, s_N\}$ denotes the set of candidate video spans, $E_Q(\cdot)$ and $E_V(\cdot)$ are frozen encoders for the query and video, respectively, and $\text{sim}(\cdot, \cdot)$ denotes a similarity function such as cosine similarity.

For instance, Moment-GPT~\cite{xu2025zero} extends the VTG-GPT~\cite{xu2024vtg} framework by combining frame-level captions from MiniGPT-v2~\cite{chen2023minigpt} with segment-level captions from Video-ChatGPT~\cite{maaz2023video}, matching them to textual queries using a similarity-based retrieval approach. TFVTG~\cite{zheng2025training} further refines this paradigm by decomposing complex queries into sub-events using an LLM, followed by segment matching through the BLIP-2 Q-Former~\cite{li2023blip}. The final predictions are derived by integrating localized spans via temporally-aware filtering that accounts for the sub-event order and relations, improving the accuracy of temporal localization.

\textbf{LLM-Driven Reasoning}: This alternative strategy treats VTG as a high-level textual inference task, leveraging the reasoning capabilities of LLMs to comprehend and localize temporal segments based on enriched video descriptions.

For example, Grounding-prompter~\cite{chen2023grounding} reformulates VTG as a long-text comprehension task, aligning speech transcriptions and visual captions with timestamp annotations. It employs a four-step multiscale denoising chain-of-thought approach, progressively refining coarse temporal predictions through iterative prompts. In a similar vein, DeVi~\cite{qin2024question} uses Video-LLaVA~\cite{lin2023video} to perform hierarchical, multi-scale captioning, followed by query-driven refinement using GPT-4o~\cite{openai2024gpt4o}. This multi-stage reasoning process allows the model to better capture event dependencies and fine-grained temporal structures, leading to more accurate localization without additional training.

Beyond these dominant strategies, emerging methods are exploring novel approaches to training-free temporal grounding. For instance, NumPro~\cite{wu2024number} introduces a unique numbering scheme inspired by manga panel sequencing, inserting numerical identifiers into each video frame to enhance temporal traceability. This subtle form of visual embedding enables LLMs to track frame sequences more effectively without modifying the model architecture or requiring fine-tuning, preserving general video comprehension while improving temporal precision.

\textbf{Summary}: Training-free VTG-MLLMs provide a lightweight, modular alternative to conventional fine-tuning approaches, effectively decomposing VTG into manageable subtasks, i.e., captioning, matching, and reasoning, without the computational overhead of extensive training. By leveraging powerful pre-trained models and off-the-shelf components, these methods reduce the cost and complexity of domain-specific adaptation, making them a compelling choice for scenarios where data availability and computational resources are limited. However, their reliance on predefined embeddings and static representations can introduce challenges in capturing fine-grained temporal dependencies, presenting an ongoing area for innovation.

\begin{figure}[t]
\centering
  \includegraphics[width=\linewidth]{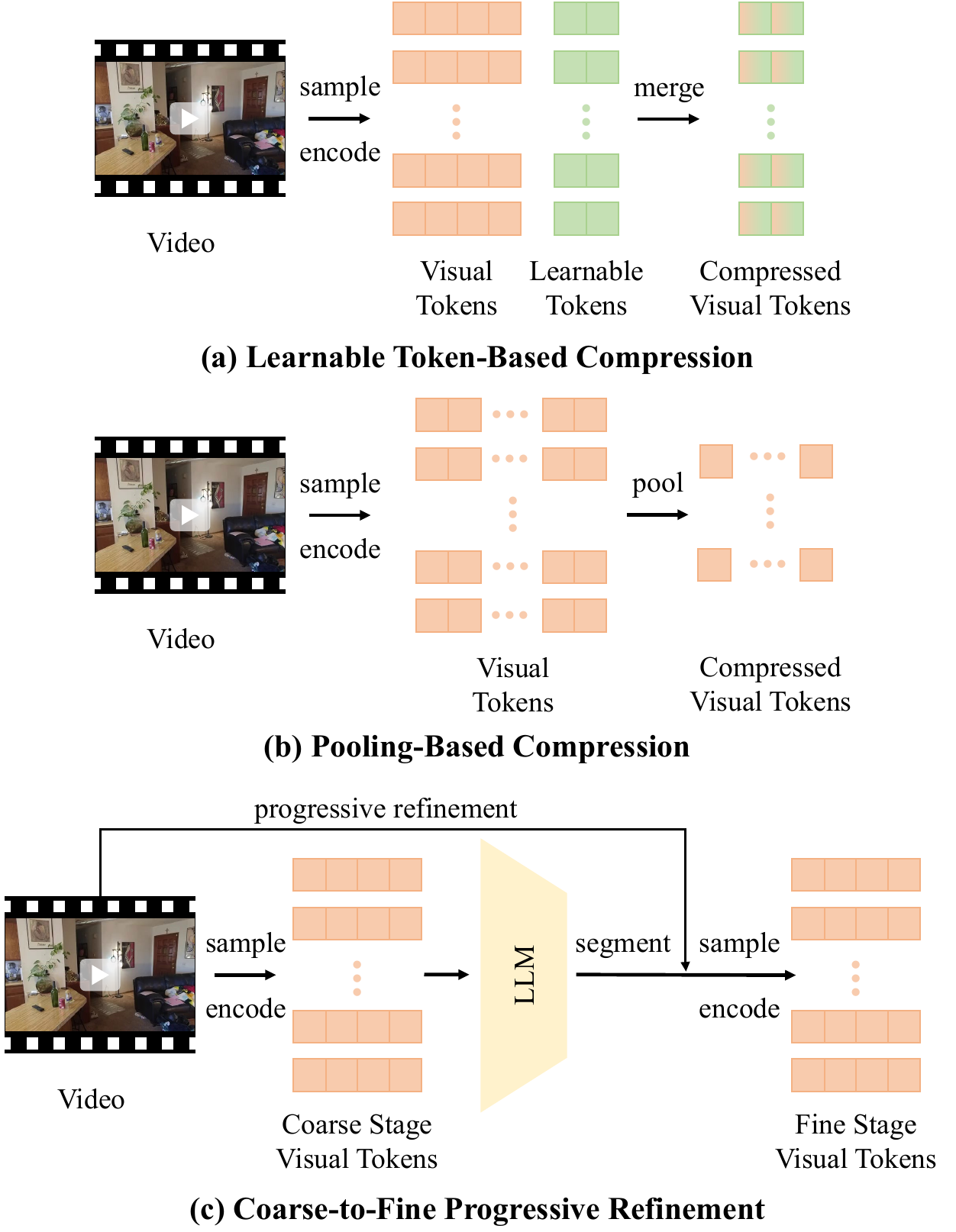}
\vspace{-0.4cm}
\caption{Representative strategies for maximizing information utility within limited input token budgets: (a) Learnable Token-Based Compression, (b) Pooling-Based Compression, and (c) Coarse-to-Fine Progressive Refinement.}
\label{Fig:compression}
\vspace{-0.6cm}
\end{figure}

\begin{figure*}[t]
\centering
    \includegraphics[width=0.867\linewidth]{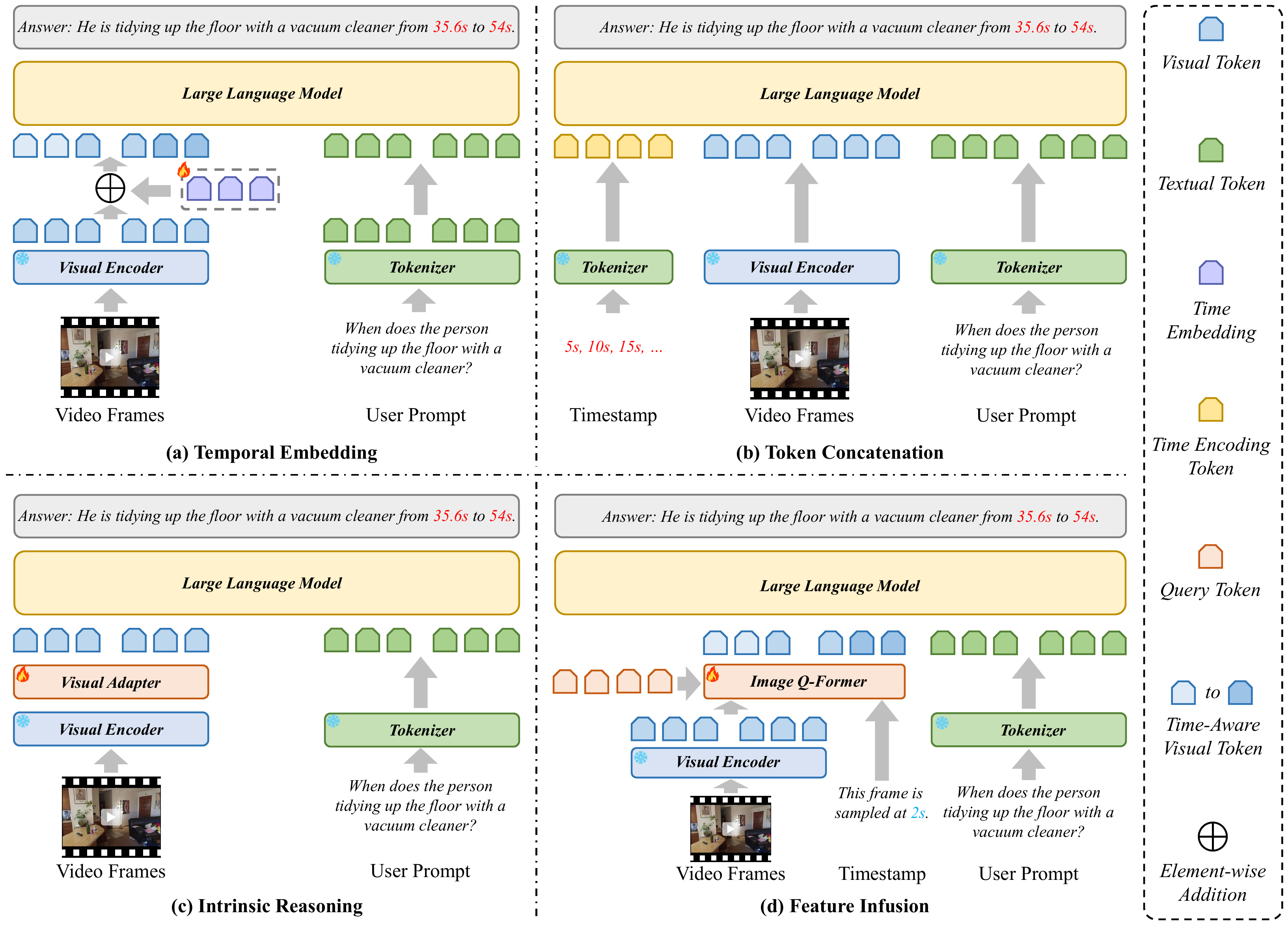}
\vspace{-0.1cm}
\caption{Illustration of temporal feature processing strategies in VTG-MLLMs, categorized into (a)-(b) \textit{Explicit Modeling}, which directly integrates timestamp information through methods like temporal embedding and token concatenation, and (c)-(d) \textit{Implicit Modeling}, which relies on the reasoning capabilities of LLMs to infer temporal relationships through intrinsic reasoning and feature infusion.}
\label{Fig:temporal feature methods}
\vspace{-0.5cm}
\end{figure*}

\subsection{Video Feature Processing in VTG-MLLMs}
\label{video feature extraction}

At the most fine-grained level of our taxonomy, we examine the video feature extraction strategies that underpin VTG-MLLMs. As discussed in the taxonomy of functional roles, Facilitator-based methods often rely on pre-trained, frozen modules to provide high-level video embeddings. In contrast, Executor-oriented designs require more sophisticated mechanisms to handle raw video inputs, reflecting their more direct involvement in temporal reasoning and event localization. This subsection focuses on the critical strategies for extracting and processing visual and temporal features within Executor-based VTG-MLLMs.

\subsubsection{Efficient Visual Feature Handling}
\label{visual feature}

Given the dense nature of frame-level information and the constrained input size of most LLMs, efficient visual feature handling is essential for capturing fine-grained temporal cues without overwhelming the model. These techniques can be broadly categorized into three main approaches, as illustrated in Fig.~\ref{Fig:compression}: \textit{Learnable Token-Based Compression}, \textit{Pooling-Based Compression}, and \textit{Coarse-to-Fine Progressive Refinement}.

\textbf{Learnable Token-Based Compression}: This strategy employs learnable tokens to compress high-dimensional visual features into a concise and manageable representation, as depicted in Fig.~\ref{Fig:compression}~(a). 
Formally, given a set of trainable tokens $\mathbf{Q}$ and raw visual features $\mathbf{V}$, the compressed representation $\mathbf{C}$ is obtained by a parameterized function $f_{\text{compress}}$:
\begin{equation} \label{eq:learnable_token_compression}
\mathbf{C} = f_{\text{compress}}(\mathbf{Q}, \mathbf{V}; \theta)
\end{equation}
where $\theta$ represents learnable parameters. 
For example, VTG-LLM~\cite{guo2024vtg} introduces Slot-Based Token Compression, where a set of learnable slot embeddings aggregate information from raw visual tokens based on similarity. TRACE~\cite{guo2024trace} adopts a similar strategy, compressing dense visual inputs into compact summaries using slot embeddings. ReVisionLLM~\cite{hannan2024revisionllm} employs a [CLS]-like token~\cite{devlin2019bert} to aggregate segment features through self-attention, providing compact yet semantically rich representations, effectively acting as another instantiation of Eqn.~(\ref{eq:learnable_token_compression}).

\textbf{Pooling-Based Compression}: Pooling techniques aggregate local or global visual features to reduce dimensionality yet retain key semantic information, as shown in Fig.~\ref{Fig:compression}~(b). For instance, LITA~\cite{huang2025lita} applies multi-granularity pooling across spatial and temporal dimensions, while Grounded-VideoLLM~\cite{wang2024grounded} and TimeMarker~\cite{chen2024timemarker} use dynamically adjustable pooling kernels to capture hierarchical visual cues. LLaVA-MR~\cite{lu2024llava} introduces token variance-based selection, prioritizing high-variance tokens to capture dynamic content more effectively. Alternatively, TimeSuite~\cite{zeng2024timesuite} reduces token overhead through token shuffling and projection, achieving compression without extra parameters.

\textbf{Coarse-to-Fine Progressive Refinement}: Rather than compressing dense frame-level features upfront, these methods progressively refine temporal predictions to improve efficiency under strict token constraints, as outlined in Fig.~\ref{Fig:compression}~(c). 
SeViLA~\cite{yu2023self} is an early example that adopts a coarse-to-fine localization strategy by selecting language-aware keyframes before answer prediction.
Similarly, HawkEye~\cite{wang2024hawkeye} adopts a recursive grounding approach, narrowing the temporal search space in iterative stages. 
Another prominent strategy involves a multi-stage refinement—first identifying a coarse temporal segment and then adjusting its boundaries—a method effectively employed by ReVisionLLM~\cite{hannan2024revisionllm}, SlowFocus~\cite{nie2024slowfocus}, and VideoMind~\cite{liu2025videomind} to reduce token overhead while preserving accuracy.

\textbf{Summary}: Efficient visual feature handling is paramount in VTG-MLLMs to reconcile the richness of dense video data with the input token limitations. The above-discussed strategies represent distinct philosophies for information reduction. Token-based and pooling methods achieve upfront compression. In contrast, progressive refinement adopts a iterative approach, selectively focusing computational resources on temporally relevant segments. Collectively, these techniques are crucial for enabling MLLMs to process detailed video sequences, striking a vital balance between capturing fine-grained temporal nuances necessary for precise grounding and maintaining computational tractability.

\subsubsection{Temporal Representation and Modeling}
\label{temporal feature subsec}
Unlike global video understanding tasks, fine-grained temporal grounding demands precise reasoning about temporal relationships to align video frames with timestamp intervals. This requirement is critical for VTG tasks such as moment retrieval and dense video captioning, which rely on accurate temporal boundary predictions. Robust timestamp representation mechanisms are essential for achieving this level of temporal precision. To address these challenges, temporal feature modeling in VTG-MLLMs can be broadly categorized into \textit{Explicit} and \textit{Implicit} modeling strategies, distinguished by whether temporal cues are directly injected into the model's input stream or contextually assimilated through its architectural design and reasoning capabilities.

\textbf{Explicit Modeling}: Explicit modeling strategies directly embed temporal information into the input or feature representations of MLLMs, as illustrated in Fig.~\ref{Fig:temporal feature methods}~(a)-(b). These approaches aim to provide precise temporal context by incorporating explicit time markers within the visual feature space, enhancing the model's ability to align video frames with timestamps. Broadly, these methods can be categorized into two main approaches: \textit{Temporal Embedding} and \textit{Token Concatenation}, each with distinct mechanisms for integrating temporal cues. 

\textit{Temporal Embedding}. One common approach involves augmenting visual tokens with dedicated temporal embeddings, effectively integrating time information into the sequence of input tokens. 
If $\mathbf{v}_{i}$ represents the $i$-th visual token and $\mathbf{e}_{t_{i}}$ is its corresponding temporal embedding, the augmented visual token $\mathbf{v}^{'}_{i}$ can be formed as:
\begin{equation} \label{eq:temporal_embedding_add}
\mathbf{v}^{'}_{i} = \mathbf{v}_{i} + \mathbf{e}_{t_{i}}.
\end{equation}
For instance, VTG-LLM~\cite{guo2024vtg} introduces learnable absolute time embeddings initialized to zero, preserving the original semantic integrity of visual tokens generated by pre-trained encoders. In contrast, LITA~\cite{huang2025lita} adopts a relative time representation, segmenting videos into $T$
equal-length chunks and assigning unique temporal tokens (e.g., $\langle 1 \rangle$ to $\langle T \rangle$) to each segment, providing a coarse but computationally efficient temporal structure. Momentor~\cite{qian2024momentor} takes a more granular approach by defining $N$ learnable anchor points, each representing a specific temporal position within the video. These anchors define a continuous temporal feature space, allowing for more precise localization through interpolation. Other methods also leverage explicit time encodings. For example, TGB~\cite{wang2024efficient}, SlowFocus~\cite{nie2024slowfocus}, and LLaVA-ST~\cite{li2025llava} incorporate temporal position embeddings to enhance temporal awareness during fine-tuning. These embeddings provide clear temporal context, improving temporal alignment without notably altering the underlying architecture of MLLMs.

\textit{Token Concatenation}. An alternative approach involves tokenizing timestamps directly from sampled frames, integrating these temporal markers with visual and textual tokens to form a unified input sequence.
If $\mathbf{S}_{P}$, $\mathbf{S}_{V}$, and $\mathbf{S}_{T}$ represent sequences of prompt tokens, visual tokens, and tokenized timestamps, respectively, the final input sequence $\mathbf{S}_{input}$ fed to the MLLM can be a concatenation:
\begin{equation} \label{eq:token_concat}
\mathbf{S}_{input} = \text{Concat}(\mathbf{S}_P, \mathbf{S}_V, \mathbf{S}_T).
\end{equation}
The specific order and interleaving strategy can vary.
For instance, LLaVA-MR~\cite{lu2024llava} dynamically selects relative frame indices or absolute timestamps based on the frame sampling rate, interleaving these markers with special tokens like $\langle time\_begin \rangle$ and $\langle time\_end \rangle$ to denote temporal boundaries. TimeMarker~\cite{chen2024timemarker} adopts a similar strategy, inserting explicit temporal separators (e.g., ``second{2.0}'') into the input sequence to enhance temporal context. Additionally, Mr.BLIP~\cite{meinardus2024surprising} systematically explores various design choices for time representation, including relative versus absolute time, decimal versus integer formats, and different token ordering schemes, evaluating their impact on model performance. TRACE~\cite{guo2024trace} further extends this approach by integrating temporal tokenization with visual feature embeddings, providing a tightly coupled representation of spatial and temporal information.

\textbf{Implicit Modeling}: Implicit modeling strategies aim to capture temporal relationships within video data through latent representations, leveraging the inherent reasoning and contextual understanding capabilities of large language models. Unlike explicit methods, which directly associate timestamps with visual inputs, implicit approaches integrate temporal cues more fluidly, embedding temporal knowledge without requiring explicit time markers. These strategies generally fall into two main categories: \textit{Intrinsic Reasoning} and \textit{Feature Infusion}, each employing distinct techniques to embed temporal context into visual representations, as illustrated in Fig.~\ref{Fig:temporal feature methods}~(c)-(d).

\textit{Intrinsic Reasoning}. This approach relies on the LLM’s inherent ability to infer temporal relationships indirectly from the interplay between visual features and time-related language prompts. Rather than embedding explicit timestamps, these methods leverage numerical cues, iterative refinement, and boundary-aware reasoning to capture temporal dynamics.
For instance, NumPro~\cite{wu2024number} introduces numerical indices directly into video frames, allowing the LLM to infer sequence order through positional awareness. Grounded-VideoLLM~\cite{wang2024grounded} adopts a similar strategy, introducing specialized temporal tokens into the LLM’s vocabulary, enabling unified modeling of time and semantics. TimeRefine~\cite{wang2024timerefine} reframes temporal grounding as a progressive refinement task, where the model first predicts coarse intervals (e.g., ``15.0s to 27.5s'') and subsequently refines these estimates by predicting offset adjustments (e.g., ``+4.0s and -1.5s''), achieving fine-grained localization through iterative reasoning.
Other models, such as VTimeLLM~\cite{huang2024vtimellm} and TPE-VLLM~\cite{li2025mitigating}, incorporate boundary-aware tasks during pretraining, explicitly teaching the model to reason about event durations and transitions, thereby enhancing temporal precision without the need for explicit time tokens.

\textit{Feature Infusion}. Feature infusion techniques integrate temporal context directly into visual feature representations by conditioning the feature extraction process. This is often achieved using architectures like Q-Formers, designed to jointly learn spatiotemporal embeddings. Formally, given raw visual features $\mathbf{V}_{raw}$ and a temporal descriptor $\mathbf{T}_{desc}$ (e.g., "This frame is sampled at 2s"), the infused features $\mathbf{F}_{infused}$ can be generated as:
\begin{equation} \label{eq:feature_infusion}
\mathbf{F}_{infused} = \text{Extractor}(\mathbf{V}_{raw}, \mathbf{T}_{desc}; \theta),
\end{equation}
where $\text{Extractor}$ (e.g., a Q-Former) processes $\mathbf{V}_{raw}$ conditioned on $\mathbf{T}_{desc}$, allowing the model to capture subtle, context-dependent temporal cues without explicit time tokenization. For example, TimeChat~\cite{ren2024timechat} and TemporalVLM~\cite{fateh2024video} leverage this by providing such temporal descriptors as conditional inputs to their Q-Former, guiding the model to incorporate temporal context. Similarly, TimeSuite~\cite{zeng2024timesuite} generates segment-level features that capture temporal dynamics across longer intervals, enabling more comprehensive temporal reasoning.

\textbf{Summary}: Temporal representation and modeling are foundational for endowing VTG-MLLMs with the capacity for precise temporal localization. Explicit modeling strategies directly furnish MLLMs with unambiguous temporal information, offering direct control and interpretability over temporal cues. Implicit modeling, on the other hand, leverages the inherent sequential processing and reasoning capabilities of LLMs or integrates temporal context more subtly during feature extraction. 
These approaches reflect ongoing exploration into how best to integrate the continuous nature of time with the discrete, symbolic processing of LLMs, ultimately shaping the model's ability to perform nuanced temporal reasoning and accurate boundary prediction.

\section{Experimental Evaluation}
\label{benchmark_experiment}
In this section, we provide a comprehensive analysis of the performance of various VTG-MLLMs across four core tasks. We begin by establishing the experimental settings, outlining the benchmark datasets and evaluation metrics that form the basis for fair comparison. Subsequently, we present a detailed performance comparison, considering both zero-shot and fine-tuning scenarios to assess their generalization capabilities and task-specific effectiveness.

\begin{table*}[t]
\caption{Comparative statistics of major datasets commonly used in VTG. These datasets span a range of tasks, including MR, DC, HD, and GQA, as described in Section \ref{tasks}. Key dataset attributes include the number of videos, average video duration, average moment duration, total number of queries, and average query length, providing a comprehensive overview of their scale and complexity.}
  \vspace{-0.1cm}
  \centering
  \label{tab:general datasets}
  \renewcommand{\arraystretch}{1.15}
  \begin{tabular}{cccccccc}
    \Xhline{0.3ex}    
    Dataset & Year & Task & \#Videos & Duration / Video & Duration / Moment & \#Queries & \#Words / Query \\
    \hline
    Charades-STA \cite{gao2017tall} & 2017 & MR & 6,672 & 0.5min & 8.1s & 16,128 & 7.2 \\
    ActivityNet-Captions \cite{krishna2017dense} & 2017 & MR \& DC & 14,926 & 1.96min & 36.2s & 71,957 & 14.8 \\
    YouCook2 \cite{zhou2018towards} & 2017 & DC & 2,000 & 5.26min & - & 15,433 & 8.8 \\
    QVHighlights \cite{lei2021detecting} & 2021 & MR \& HD & 10,148 & 2.5min & 24.6s & 10,310 & 11.3 \\
    NExT-GQA \cite{xiao2024can} & 2023 & GQA & 1,557 & 0.70min & 6.92s & 8,911 & - \\
    \Xhline{0.3ex}
  \end{tabular}
  \vspace{-0.5cm}
\end{table*}

\subsection{Experimental Settings}
\label{sec:experimental_settings}
This subsection outlines the fundamental experimental settings employed in this survey. We first introduce the \textit{Benchmark Datasets} used for evaluation, followed by a description of the \textit{Evaluation Metrics} employed to assess model performance across different VTG tasks.

\subsubsection{Benchmark Datasets}
\label{sec: benchmark datasets}
Benchmark datasets are foundational for the development and rigorous evaluation of VTG research, providing essential video content and corresponding natural language annotations necessary for model training and standardized comparisons. The current VTG dataset landscape can be broadly categorized into two groups: \textit{Standard Benchmarks}, which are well-established and widely adopted by the research community, and \textit{Emerging Benchmarks}, designed to address gaps in existing datasets by targeting complex scenarios such as long-term temporal reasoning and fine-grained event distinctions.

Given space limitations, we focus this subsection on key characteristics of select standard benchmarks that are utilized in our experimental evaluations (Sections~\ref{sec:zero_shot_performance} and \ref{sec:fine_tuning_performance}). Comparative statistics for these datasets are summarized in Table~\ref{tab:general datasets}. A comprehensive review, including additional standard and emerging benchmarks, is provided in the supplementary material. 

\textbf{Charades-STA}~\cite{gao2017tall},  derived from the Charades dataset~\cite{sigurdsson2016hollywood}, is widely employed for MR. It includes approximately 6.7k videos depicting indoor daily activities, annotated with about 16.1k query-moment pairs. Videos average 30 seconds, moments average 8 seconds, and textual queries are concise (7.2 words on average), making it suitable for action-centric grounding tasks.

\textbf{ActivityNet-Captions}~\cite{krishna2017dense},  an extension of the human activity dataset ActivityNet~\cite{caba2015activitynet}, is primarily utilized for DC, though also popular in MR tasks. It features around 15k videos annotated with multiple temporally localized events per video (avg. 3.65 events per video). With an average video duration of 1.96 minutes, average moment length of 36.2 seconds, and longer descriptive queries (14.8 words), making it suitable for evaluating models on complex temporal structures and narratives.

\textbf{YouCook2}~\cite{zhou2018towards} offers a large-scale instructional video dataset tailored for procedural activity understanding, primarily applied in DC. It comprises 2,000 long, untrimmed cooking videos, collectively spanning 176 hours and covering 89 distinct recipes, resulting in approximately 15.4k query-moment pairs. Each video is densely annotated with precise temporal boundaries for step-level imperative sentences (avg. 8.8 words).

\textbf{QVHighlights}~\cite{lei2021detecting} is specifically designed for query-based HD and also finds application in MR. The dataset includes approximately 10.1k YouTube videos, each cropped to a fixed duration of 150 seconds, and features around 10.3k query-moment pairs. A distinctive aspect of QVHighlights is its allowance for multiple disjoint moments per query, reflecting the non-linear nature of salient video highlights, which average about 24.6 seconds. Evaluation of its test set is managed via an official server\footnote{Evaluation: \url{https://codalab.lisn.upsaclay.fr/competitions/6937}}.

\textbf{NExT-GQA}~\cite{xiao2024can} extends the NExT-QA dataset~\cite{xiao2021next}, designed to support research on QA tasks that require temporal grounding as evidence. It adds precise temporal annotations for approximately 8.9k QA pairs across 1.6k videos, pinpointing critical segments (avg. 6.9 seconds) needed to answer questions. A notable feature is its weakly-supervised setup where only validation and test sets provide ground-truth temporal labels.

\begin{table*}[t]
    \centering
    \caption{Zero-shot performance comparison on video moment retrieval benchmarks (Charades-STA~\cite{gao2017tall} and ActivityNet-Captions~\cite{krishna2017dense}). Methods are grouped by training paradigm (PT: pretraining, TF: training-free). The best results are \textbf{boldfaced}, and the second-best results are \underline{underlined}. This formatting convention is uniformly adopted across all subsequent tables unless explicitly noted otherwise.}
    \label{tab:charades_activitynet}
    \vspace{-0.1cm}
    \renewcommand{\arraystretch}{1.17}
    \begin{tabular}{l|c|cccc|cccc}
        \Xhline{0.3ex}
        \multirow{2}{*}{Method} & \multirow{2}{*}{Paradigm} & \multicolumn{4}{c|}{Charades-STA \cite{gao2017tall}} & \multicolumn{4}{c}{ActivityNet-Captions \cite{krishna2017dense}} \\
        & & mIoU & R@0.3 & R@0.5 & R@0.7 & mIoU & R@0.3 & R@0.5 & R@0.7 \\
        \hline
        Time-R1 \cite{wang2025timezero} & PT & - & \textbf{78.1} & \textbf{60.8} & \textbf{35.3} & - & - & - & - \\
        VideoMind \cite{liu2025videomind} & PT & \textbf{50.2} & \underline{73.5} & \underline{59.1} & \underline{31.2} & 33.3 & 48.4 & 30.3 & 15.7 \\
        TimeMarker \cite{chen2024timemarker} & PT & \underline{48.4} & \underline{73.5} & 51.9 & 26.9 & - & - & - & - \\
        VideoChat-T \cite{zeng2024timesuite} & PT & - & 69.9 & 48.7 & 24.0 & - & - & - & - \\
        LLaVA-ST \cite{li2025llava} & PT & 42.4 & 63.1 & 44.8 & 23.4 & - & - & - & - \\
        MeCo \cite{pang2025measure} & PT & - & - & 44.4 & 17.5 & - & - & - & - \\
        E.T.Chat \cite{liu2024bench} & PT & - & - & 43.2 & 19.4 & - & - & - & - \\
        VideoExpert \cite{zhao2025videoexpert} & PT & 41.1 & 61.5 & 40.3 & 20.9 & - & - & - & - \\
        TRACE \cite{guo2024trace} & PT & - & - & 40.3 & 19.4 & \textbf{39.0} & - & \textbf{37.7} & \textbf{24.0} \\
        VideoChat-TPO \cite{yan2024task} & PT & 38.1 & 58.3 & 40.2 & 18.4 & - & - & - & - \\
        MLLM-TA \cite{liu2024mllm} & PT & - & - & 37.9 & 18.1 & - & - & 27.6 & 18.3 \\
        Ground-VideoLLM \cite{wang2024grounded} & PT & 36.8 & 54.2 & 36.4 & 19.7 & \underline{36.1} & 46.2 & 30.3 & 19.0 \\
        VTG-LLM \cite{guo2024vtg} & PT & - & - & 33.8 & 15.7 & - & - & - & - \\
        TPE-VLLM \cite{li2025mitigating} & PT & 34.7 & 55.5 & 33.1 & 14.7 & \underline{36.1} & \textbf{50.4} & \underline{35.4} & \underline{19.2} \\
        TimeChat \cite{ren2024timechat} & PT & - & - & 32.2 & 13.4 & - & - & - & - \\
        HawkEye \cite{wang2024hawkeye} & PT & 33.7 & 50.6 & 31.4 & 14.5 & 32.7 & \underline{49.1} & 29.3 & 10.7 \\
        TemporalVLM \cite{fateh2024video} & PT & - & - & 30.1 & 13.2 & - & - & - & - \\
        GroundingGPT \cite{li2024groundinggpt} & PT & - & - & 29.6 & 11.9 & - & - & - & - \\
        VTimeLLM \cite{huang2024vtimellm} & PT & 31.2 & 51.0 & 27.5 & 11.4 & 30.4 & 44.0 & 27.8 & 14.3 \\
        Momentor \cite{qian2024momentor} & PT & 28.5 & 42.6 & 26.6 & 11.6 & 29.3 & 42.9 & 23.0 & 12.4 \\
        \hline
        TFVTG \cite{zheng2025training} & TF & \textbf{44.5} & \textbf{67.0} & \textbf{50.0} & \underline{24.3} & \textbf{34.1} & \underline{49.3} & 27.0 & \underline{13.4} \\
        VTG-GPT \cite{xu2024vtg} & TF & \underline{39.8} & \underline{59.5} & \underline{43.7} & \textbf{25.9} & 30.5 & 47.1 & \underline{28.3} & 12.8 \\
        Moment-GPT \cite{xu2025zero} & TF & 36.5 & 58.2 & 38.4 & 21.6 & \underline{30.8} & \textbf{48.1} & \textbf{31.1} & \textbf{14.9} \\
        ChatVTG \cite{qu2024chatvtg} & TF & 34.9 & 52.7 & 33.0 & 15.9 & 27.2 & 40.7 & 22.5 & 9.4 \\
        \Xhline{0.3ex}
  \end{tabular}
  \vspace{-0.5cm}
\end{table*}

\subsubsection{Evaluation Metrics}

Effective evaluation of VTG-MLLMs relies on well-defined metrics that assess both the temporal localization accuracy and semantic alignment of predicted moments with ground-truth segments. 

\textbf{Video Moment Retrieval} is quantified by two key metrics for localization accuracy: 
\textit{Mean Intersection over Union (mIoU)}~\cite{gao2017tall} quantifies the average overlap between predicted and ground-truth segments, indicating localization accuracy; and
\textit{Recall at Rank $n$ with an IoU threshold of $m$ (R@$n$ (IoU=$m$))}~\cite{hu2016natural} quantifies the percentage of cases where the top-$n$ retrieved segments have an IoU with the ground truth greater than or equal to the threshold $m$, under varying IoU thresholds (e.g., R@1 at IoU=0.3/0.5/0.7).

\textbf{Dense Video Captioning} hinges on both temporal precision and linguistic quality. Key metrics include: 
\textit{SODA\_c}\cite{fujita2020soda} assesses structural alignment between captions and temporal segments, emphasizing event segmentation coherence; \textit{METEOR}\cite{banerjee2005meteor} and \textit{CIDEr}~\cite{vedantam2015cider} evaluate linguistic similarity between generated and reference captions, considering language precision and recall; and \textit{F1 Score} measures the harmonic mean of precision and recall for instances within a video, assessing coverage and relevance.

\textbf{Video Highlight Detection} commonly relies on metrics:
\textit{Mean Average Precision (mAP)} measures the area under the precision-recall curve at various IoU thresholds, such as 0.5 and 0.75, reflecting the model's ability to accurately rank and localize key moments; and \textit{HIT@1}~\cite{liu2015multi} calculates the proportion of instances where the top-ranked highlight prediction matches a ground-truth highlight, providing a straightforward measure of first-choice accuracy.

\textbf{Temporally Grounded Video Question Answering} synthesizes metrics for both temporal localization and question-answering accuracy:
\textit{mIoU} follows the same setting as in MR; \textit{Mean Intersection over Prediction (mIoP)}\cite{xiao2024can} evaluates the proportion of the predicted segment overlapping with the ground-truth moment, as an alternative alignment metric;
\textit{Acc@QA}\cite{zhong2022video} measures the percentage of correctly answered questions, irrespective of temporal grounding; and \textit{Acc@GQA}~\cite{xiao2024can} extends Acc@QA by requiring correct answers with accurate temporal grounding, where the predicted segment must achieve an IoP score of at least 0.5.

\subsection{Zero-Shot Performance Comparison}
\label{sec:zero_shot_performance}
Zero-shot evaluation measures the ability of VTG-MLLMs to generalize to new datasets without any dataset-specific fine-tuning. This evaluation is crucial for understanding the inherent adaptability and robustness of different models, reflecting their capacity to transfer knowledge across diverse video grounding tasks.
For clarity, models are categorized based on their training paradigms: \textit{Pretraining (PT)} approaches (Section~\ref{pretraining subsec}), which rely on large-scale multimodal datasets to build general-purpose grounding capabilities, and \textit{Training-Free (TF)} approaches (Section~\ref{training-free subsec}), which leverage pre-trained foundation models without requiring additional task-specific training.

\textbf{Video Moment Retrieval}. Table~\ref{tab:charades_activitynet} presents a comparison of zero-shot performance on the Charades-STA~\cite{gao2017tall} and ActivityNet-Captions~\cite{krishna2017dense} benchmarks. 

\textit{PT Approaches}: Among pretraining-based methods, models like Time-R1~\cite{wang2025timezero}, VideoMind~\cite{liu2025videomind}, and TimeMarker~\cite{chen2024timemarker} demonstrate state-of-the-art zero-shot performance, particularly on the Charades-STA dataset. 
Time-R1, in particular, success in achieving the highest recall scores likely stems from its innovative use of reinforcement learning.
Similarly, TRACE~\cite{guo2024trace} and TPE-VLLM~\cite{li2025mitigating} achieve competitive results on ActivityNet-Captions, benefiting from their fine-grained temporal modeling and boundary-aware pertaining.

\textit{TF Approaches}: In the training-free category, methods like TFVTG~\cite{zheng2025training} and VTG-GPT~\cite{xu2024vtg} also exhibit competitive performance, particularly on Charades-STA. Notably, TFVTG outperforms several pretraining-based models in terms of mIoU and R@0.5. These models benefit from flexible, modular designs that leverage powerful pre-trained language and vision models without requiring task-specific fine-tuning.

\begin{table*}[t]
    \centering
    \caption{Zero-Shot performances on dense video captioning benchmarks (ActivityNet-Captions \cite{krishna2017dense} and YouCook2 \cite{zhou2018towards}).}
    \vspace{-0.1cm}
    \label{tab:dense_captioning}
    \renewcommand{\arraystretch}{1.15}
    \begin{tabular}{l|c|cccc|ccc}
        \Xhline{0.3ex}
        \multirow{2}{*}{Method} & \multirow{2}{*}{Paradigm} & \multicolumn{4}{c|}{ActivityNet-Captions \cite{krishna2017dense}} & \multicolumn{3}{c}{YouCook2 \cite{zhou2018towards}} \\
        % \cline{2-8}
        & & SODA\_c & CIDEr & METEOR & F1 Score & SODA\_c & CIDEr & F1 Score \\
        \hline
        TRACE \cite{guo2024trace} & PT & \textbf{6.0} & \underline{25.9} & \underline{6.4} & \textbf{39.3} & \textbf{2.2} & \textbf{8.1} & \textbf{22.4} \\
        Ground-VideoLLM \cite{wang2024grounded} & PT & \textbf{6.0} & - & \textbf{6.8} & - & - & - & - \\
        VTimeLLM \cite{huang2024vtimellm} & PT & \underline{5.8} & \textbf{27.6} & \textbf{6.8} & - & 0.9 & 3.4 & - \\
        VideoExpert \cite{zhao2025videoexpert} & PT & - & - & - & - & \underline{2.1} & \underline{6.0} & - \\
        VTG-LLM \cite{guo2024vtg} & PT & 5.1 & 20.7 & 5.9 & 34.8 & 1.5 & 5.0 & \underline{17.5} \\
        TimeChat \cite{ren2024timechat} & PT & 4.7 & 19.0 & 5.7 & \underline{36.9} & 1.2 & 3.4 & 12.6 \\
        Momentor \cite{qian2024momentor} & PT & 2.3 & 14.9 & 4.7 & - & - & - & - \\
        TemporalVLM \cite{fateh2024video} & PT & - & - & - & - & 1.2 & 3.7 & 13.1 \\
        \Xhline{0.3ex}
    \end{tabular}
    \vspace{-0.4cm}
\end{table*}

\textbf{Dense Video Captioning}. 
Table~\ref{tab:dense_captioning} presents zero-shot performance comparisons on the ActivityNet-Captions~\cite{krishna2017dense} and YouCook2~\cite{zhou2018towards} benchmarks. 

\textit{PT Approaches}: Within the pretraining category, methods like TRACE~\cite{guo2024trace}, Grounded-VideoLLM~\cite{wang2024grounded}, and VTimeLLM~\cite{huang2024vtimellm} consistently achieve strong results across key metrics, including SODA\_c, CIDEr, and METEOR. TRACE stands out as a top performer on both ActivityNet-Captions and YouCook2, reflecting its effective integration of fine-grained temporal encoding and sequence generation. 

\textit{TF Approaches}: Training-free methods are generally not included in this evaluation, as they often rely on offline MLLMs for generating descriptions, which do not directly align with the standard online evaluation protocols typically used for dense captioning.

\textbf{Video Highlight Detection}. 
Table~\ref{tab:qvhighlights_zs} presents zero-shot performance comparisons on the QVHighlights~\cite{lei2021detecting} benchmark, a challenging dataset specifically designed for localizing query-relevant highlight moments within videos. 

\textit{PT Approaches}: Among pretraining-based methods, VideoExpert~\cite{zhao2025videoexpert} stands out with the highest mAP scores, reflecting its strong ability to capture fine-grained temporal relationships and accurately rank highlight moments. In contrast, videoChat-T~\cite{zeng2024timesuite} leads in Hit@1, demonstrating superior single-shot localization accuracy, which highlights its effectiveness in identifying salient temporal regions.

\textit{TF Approaches}: Training-free methods are typically excluded from direct comparison in this context, as they often rely on frame-query similarity scoring, which inherently aligns well with highlight detection tasks. These approaches, while effective, lack the structured, end-to-end temporal grounding capabilities required for competitive performance in strictly defined zero-shot scenarios.

\begin{table}[t]
    \centering
    \caption{Zero-Shot performance comparison on video highlight detection benchmark QVHighlights \cite{lei2021detecting}.}
    \label{tab:qvhighlights_zs}
    \vspace{-0.1cm}
    \renewcommand{\arraystretch}{1.15}
    \begin{tabular}{l|c|cc}
        \Xhline{0.3ex}
        Method & Paradigm & mAP & HIT@1 \\
        \hline
        VideoExpert \cite{zhao2025videoexpert} & PT & \textbf{35.8} & \underline{52.7} \\
        TRACE \cite{guo2024trace} & PT & \underline{26.8} & 42.7 \\
        VideoChat-T \cite{zeng2024timesuite} & PT & 26.5 & \textbf{54.1} \\
        MLLM-TA \cite{liu2024mllm} & PT & 19.2 & 31.8 \\
        VTG-LLM \cite{guo2024vtg} & PT & 16.5 & 33.5 \\
        TemporalVLM \cite{fateh2024video} & PT & 16.4 & 31.3 \\
        TimeChat \cite{ren2024timechat} & PT & 14.5 & 23.9 \\
        Momentor \cite{qian2024momentor} & PT & 7.6 & - \\
        \Xhline{0.3ex}
    \end{tabular}
    \vspace{-0.5cm}
\end{table}

\begin{table*}[t]
    \centering
    \caption{Zero-Shot performance comparison on temporally grounded video QA benchmark NExT-GQA \cite{xiao2024can}.}
    \label{tab:next_gqa}
    \vspace{-0.1cm}
    \renewcommand{\arraystretch}{1.15}
    \begin{tabular}{lccccccccc}
        \Xhline{0.3ex}
        Method & Paradigm & mIoP & IoP@0.3 & IoP@0.5 & mIoU & IoU@0.3 & IoU@0.5 & Acc@GQA & Acc@QA \\
        \hline
        DeVi \cite{qin2024question} & TF & \textbf{39.3} & - & \textbf{37.9} & 22.3 & - & 17.4 & \underline{28.0} & \textbf{71.6} \\
        VideoMind \cite{liu2025videomind} & PT & \underline{39.0} & \textbf{56.0} & \underline{35.3} & \textbf{31.4} & \textbf{50.2} & \textbf{25.8} & \textbf{28.2} & - \\
        VideoChat-TPO \cite{yan2024task} & PT & 35.6 & \underline{47.5} & 32.8 & 27.7 & \underline{41.2} & \underline{23.4} & 25.5 & - \\
        VideoExpert \cite{zhao2025videoexpert} & PT & 34.6 & 45.3 & 29.3 & \underline{27.9} & 41.0 & 22.4 & 21.6 & \underline{71.1} \\
        HawkEye \cite{wang2024hawkeye} & PT & - & - & - & 25.7 & 37.0 & 19.5 & - & - \\
        Ground-VideoLLM \cite{wang2024grounded} & PT & 34.5 & 42.6 & 34.4 & 21.1 & 30.2 & 18.0 & 26.7 & - \\
        SeViLA \cite{yu2023self} & PT & 29.5 & 34.7 & 22.9 & 21.7 & 29.2 & 13.8 & 16.6 & 68.1 \\
        VTimeLLM \cite{huang2024vtimellm} & PT & 27.9 & 30.3 & 23.8 & 18.3 & 27.7 & 14.1 & 12.7 & - \\
        \Xhline{0.3ex}
    \end{tabular}
    \vspace{-0.4cm}
\end{table*}

\textbf{Temporally Grounded Video Question Answering}. 
Table~\ref{tab:next_gqa} presents zero-shot performance on the NExT-GQA~\cite{xiao2024can} benchmark, a relatively recent dataset designed to evaluate temporally grounded question answering in videos. 

\textit{PT Approaches}: Among pretraining-based methods, VideoMind~\cite{liu2025videomind} showcases exceptional strength in temporal localization, achieving the highest scores across all IoU-related metrics. This proficiency is complemented by its outstanding Acc@GQA performance, largely attributed to its methodical and step-by-step approach to solving GQA problems. In parallel, VideoChat-TPO~\cite{yan2024task} demonstrates robust and competitive results, underscoring the efficacy of its unique task preference optimization strategy.

\textit{TF Approaches}: In the training-free category, DeVi~\cite{qin2024question} stands out, achieving the highest scores in mIoP, Acc@GQA, and Acc@QA. 
This indicates a strong capacity for fine-grained video temporal understanding, likely due to the carefully designed pipeline that effectively combines the perceptual capabilities of the pretrained MLLMs with the reasoning ability of LLMs.

\textbf{Summary}: Experimental results across the four tasks suggest that \textit{PT Approaches} generally exhibit strong zero-shot performance due to their explicit temporal learning and large-scale data exposure, particularly excelling in tasks like moment retrieval. While \textit{TF Approaches} are not directly comparable in dense captioning and highlight detection due to architectural differences, models like DeVi~\cite{qin2024question} show that with well-designed pipelines, \textit{TF Approaches} can still achieve strong temporal reasoning in suitable tasks.

\begin{table}[t]
    \centering
    \caption{Fine-Tuning performance comparison on video moment retrieval benchmark Charades-STA \cite{gao2017tall}.}
    \label{tab:video_moment_retrieval_ft}
    \vspace{-0.1cm}
    \renewcommand{\arraystretch}{1.15}
    \begin{tabular}{l|c|ccc}
        \Xhline{0.3ex}
        \multicolumn{5}{c}{\textbf{Charades-STA \cite{gao2017tall}}} \\
        \hline
        Method & Paradigm & mIoU & R@0.5 & R@0.7 \\
        \hline
        LLaVA-MR \cite{lu2024llava} & FT & \textbf{59.8} & \textbf{70.7} & \textbf{49.6} \\
        Mr.BLIP \cite{meinardus2024surprising} & FT & \underline{58.6} & \underline{69.3} & \underline{49.3} \\
        VideoLights \cite{paul2024videolights} & FT & 52.9 & 62.0 & 41.1 \\
        ReCorrect \cite{bao2024vid} & FT & 48.7 & 54.4 & 31.1 \\
        LMR \cite{liu2024context} & FT & - & 55.9 & 35.2 \\
        LLaViLo \cite{ma2023llavilo} & FT & - & 55.7 & 33.4 \\
        \hline
        Time-R1 \cite{wang2025timezero} & PT & - & \textbf{72.2} & \underline{50.1} \\
        VideoChat-R1 \cite{li2025videochat} & PT & \textbf{60.8} & \underline{71.7} & \textbf{50.2} \\
        VideoChat-T \cite{zeng2024timesuite} & PT & - & 67.1 & 43.0 \\
        VideoChat-TPO \cite{yan2024task} & PT & \underline{55.0} & 65.0 & 40.7 \\
        SpaceVLLM \cite{wang2025spacevllm} & PT & - & 63.6 & 38.5 \\
        TRACE \cite{guo2024trace} & PT & - & 61.7 & 41.4 \\
        VideoExpert \cite{zhao2025videoexpert} & PT & 52.2 & 60.8 & 36.5 \\
        HawkEye \cite{wang2024hawkeye} & PT & 49.3 & 58.3 & 28.8 \\
        VTG-LLM \cite{guo2024vtg} & PT & - & 57.2 & 33.4 \\
        TemporalVLM \cite{fateh2024video} & PT & - & 54.5 & 29.0 \\
        MLLM-TA \cite{liu2024mllm} & PT & - & 48.9 & 25.3 \\
        TimeChat \cite{ren2024timechat} & PT & - & 46.7 & 23.7 \\
        \Xhline{0.3ex}
    \end{tabular}
    \vspace{-0.2cm}
\end{table}

\begin{table}[t]
    \centering
    \caption{Fine-Tuning performance comparison on dense video captioning benchmark YouCook2 \cite{zhou2018towards}.}
    \label{tab:youcook2_ft}
    \vspace{-0.1cm}
    \renewcommand{\arraystretch}{1.15}
    \begin{tabular}{l|c|ccc}
        \Xhline{0.3ex}
        Method & Paradigm & SODA\_c & CIDEr & F1 Score \\
        \hline
        TRACE \cite{guo2024trace} & PT & \textbf{6.7} & \textbf{35.5} & \textbf{31.8} \\
        VideoExpert \cite{zhao2025videoexpert} & PT & \underline{4.2} & \underline{18.7} & - \\
        VTG-LLM \cite{guo2024vtg} & PT & 3.6 & 13.4 & \underline{20.6} \\
        TimeChat \cite{ren2024timechat} & PT & 3.4 & 11.0 & 19.5 \\
        TemporalVLM \cite{fateh2024video} & PT & 3.4 & 13.2 & 20.0 \\
        \Xhline{0.3ex}
    \end{tabular}
    \vspace{-0.5cm}
\end{table}

\begin{table}[t]
    \centering
    \caption{Fine-Tuning performance comparison on video highlight detection benchmark QVHighlights \cite{lei2021detecting}.}
    \label{tab:qvhighlights_ft} 
    \vspace{-0.1cm}
    \renewcommand{\arraystretch}{1.15}
    \begin{tabular}{l|c|cc}
        \Xhline{0.3ex}
        Method & Paradigm & mAP & HIT@1 \\
        \hline
        VideoChat-TPO \cite{yan2024task} & PT & \textbf{38.8} & \textbf{66.2} \\
        VideoExpert \cite{zhao2025videoexpert} & PT & \underline{36.1} & \underline{61.0} \\
        TRACE \cite{guo2024trace} & PT & 31.8 & 51.5 \\
        VideoChat-T \cite{zeng2024timesuite} & PT & 27.0 & 55.3 \\
        TemporalVLM \cite{fateh2024video} & PT & 25.1 & 43.0 \\
        VTG-LLM \cite{guo2024vtg} & PT & 24.1 & 41.3 \\
        MLLM-TA \cite{liu2024mllm} & PT & 23.9 & 40.1 \\
        TimeChat \cite{ren2024timechat} & PT & 21.7 & 37.9 \\
        \Xhline{0.3ex}
    \end{tabular}
    \vspace{-0.5cm}
\end{table}

\subsection{Fine-Tuning Performance Comparison}
\label{sec:fine_tuning_performance}
Fine-tuning evaluation assesses the performance of VTG-MLLMs after they are trained on the train-split of specific benchmark datasets. Unlike zero-shot evaluation, which tests the generalization ability of models without exposure to task-specific data, this setting allows models to adapt to domain-specific distributions and task requirements, often resulting in substantial performance gains. 
Notably, the NExT-GQA~\cite{xiao2024can} benchmark is excluded from this evaluation, as its dataset definition does not provide a dedicated training split, precluding fine-tuning.

To better reflect differences in training paradigms, we categorize models into two types: \textit{Pretraining (PT)} (Section~\ref{pretraining subsec}) and \textit{Fine-Tuning (FT)} (Section~\ref{fine-tuning subsection}). The \textit{PT} refers to models that undergo large-scale multimodal pretraining prior to task-specific fine-tuning on downstream training splits. In contrast, the \textit{FT} is specifically designed to be trained directly on the training split of the downstream task.

\textbf{Video Moment Retrieval}. Table~\ref{tab:video_moment_retrieval_ft} presents the fine-tuning performance on Charades-STA~\cite{gao2017tall}.

\textit{PT Approaches}: Pretraining-based models exhibit significant performance enhancements when fine-tuned on the benchmark. Notably, reinforcement learning-based methods, Time-R1~\cite{wang2025timezero} and VideoChat-R1~\cite{li2025videochat}, emerge as the top performers, underscoring the profound potential of RL in the MLLM fine-tuning process. Closely following these leaders, models like VideoChat-T~\cite{zeng2024timesuite} and VideoChat-TPO~\cite{yan2024task} also deliver highly competitive outcomes. These results underscore the benefit of large-scale pretraining as a foundation for task-specific adaptation, allowing models to further refine their temporal grounding capabilities.

\textit{FT Approaches}: Models designed for direct fine-tuning also exhibit remarkable performance, particularly on Charades-STA. LLaVA-MR~\cite{lu2024llava} leads with the highest mIoU, R@0.5, and R@0.7, outperforming all PT methods on these metrics. Mr.BLIP~\cite{meinardus2024surprising} is a close second with mIoU and strong recall scores. On ActivityNet-Captions, Mr.BLIP~\cite{meinardus2024surprising} achieves the best R@0.5 and R@0.7 across all reported models. This demonstrates that well-optimized fine-tuning can effectively exploit the strong generalization capabilities of existing large vision-language models without requiring extensive pretraining related to temporal understanding.

\textbf{Dense Video Captioning} and \textbf{Video Highlight Detection}. The fine-tuning performance for DC on YouCook2~\cite{zhou2018towards} and HD on QVHighlights~\cite{lei2021detecting} are presented in Table~\ref{tab:youcook2_ft} and Table~\ref{tab:qvhighlights_ft}, respectively.

\textit{PT Approaches}:
For DC on YouCook2 , TRACE~\cite{guo2024trace} stands out significantly among pretraining-based methods, achieving the highest SODA\_c, CIDEr, and F1 Score. VideoExpert~\cite{zhao2025videoexpert} also shows competitive results, particularly in SODA\_c and CIDEr. 
Turning to HD on QVHighlights, VideoChat-TPO~\cite{yan2024task} demonstrates exceptional performance among PT approaches, leading with the highest mAP and Hit@1. VideoExpert~\cite{zhao2025videoexpert} also achieves a competitive mAP, while VideoChat-T~\cite{zeng2024timesuite} shows a strong Hit@1 score.
For both DC and HD, these results show that PT models can be effectively adapted to the specific nuances of these tasks through fine-tuning.

\textit{FT Approaches}:
For both DC and HD, results for \textit{FT approaches} are absent in the presented tables under standard fine-tuning protocols. This is primarily due to two factors. Firstly, \textit{FT approaches} of the "Offline Textualization with MLLMs" type (as described in Section~\ref{fine-tuning subsection}) adopt architectures incompatible with the evaluation settings of DC and HD, which is analogous to the exclusion of \textit{TF approaches} from zero-shot evaluations on these benchmarks.
Secondly, "Direct Fine-Tuning of MLLMs" for these specific tasks, while theoretically possible, remains unreported in current public research efforts. The complexity of DC (generating multiple, temporally ordered descriptions) and the nuanced requirements of HD (identifying subtle, query-relevant highlights) may necessitate more specialized architectural adaptations or fine-tuning strategies for direct application of MLLMs, which are less explored in current literature compared to their application in MR. 

\textbf{Summary}: Fine-tuning enables VTG-MLLMs to better align with task-specific objectives and dataset characteristics, leading to notable performance gains. \textit{PT Approaches} like TRACE~\cite{guo2024trace} and VideoExpert~\cite{zhao2025videoexpert} benefit significantly from this process, consistently ranking among top performers across tasks. Meanwhile, \textit{FT Approaches} such as LLaVA-MR~\cite{lu2024llava} and Mr.BLIP~\cite{meinardus2024surprising} achieve competitive or even superior results in MR, highlighting the effectiveness of direct supervision. The absence of \textit{FT Approaches} in tasks like DC and HD reflects current architectural limitations, marking this as an underexplored yet promising direction for future work.

\section{Limitations and Future Directions}
\label{future}
The integration of LLMs into VTG has led to significant advancements, showing remarkable potential in understanding and localizing temporal events within videos. However, several critical limitations remain.
This section outlines key challenges and proposes future research directions, focusing on 1) training paradigms, 2) feature representation, 3) temporal modeling, and 4) multimodal integration.

\textbf{Training Paradigms}: 
While fine-tuning (Section~\ref{fine-tuning subsection}) and training-free (Section~\ref{training-free subsec}) approaches offer practical advantages, their performance is fundamentally capped by the capabilities inherited from their pretrained foundations~\cite{lin2023video, maaz2023video}. These strategies are inherently constrained in their ability to instill the fine-grained temporal nuances essential for complex VTG, particularly if such understanding is lacking in the base model.

To address this limitation, future research should advance along two complementary paths. 
The primary path involves enhancing the pretraining paradigm (Section~\ref{pretraining subsec}) itself, by developing more targeted objectives and datasets that foster a deeper, native temporal perception in foundation models.
A parallel, more agile path lies in pioneering advanced post-training techniques such as RL, offering a powerful mechanism to sharpen a model's specialized skills beyond the reach of conventional Supervised Fine-Tuning (SFT). 
The synergy between these techniques
will likely define the next generation of state-of-the-art VTG-MLLMs.

\textbf{Feature Representation}: Efficient and effective feature representation remains a significant challenge in VTG-MLLMs. Accurate temporal grounding, particularly in long videos or tasks requiring fine-grained event distinction, demands the processing of dense, high-resolution visual inputs. However, this directly conflicts with the strict input token limitations of current LLM architectures. Existing feature compression strategies (Section~\ref{visual feature}) help mitigate this issue but often risk discarding critical temporal cues, potentially compromising grounding accuracy.

To overcome this, future work could explore adaptive token selection strategies that dynamically prioritize the most relevant visual tokens
% based on contextual importance,
while aggressively filtering out redundant information. Techniques such as temporal saliency detection, attention-based token pruning, and hierarchical feature aggregation could significantly reduce the computational burden without sacrificing precision. Moreover, developing more compact yet semantically rich video embeddings could further enhance scalability across a broad range of video understanding tasks~\cite{wang2024retake, li2024llama}.

\textbf{Temporal Modeling}: Accurate temporal reasoning is central to VTG but remains an open challenge. Current approaches vary widely in their representation of temporal information, including explicit (e.g., absolute timestamps~\cite{chen2024timemarker}, frame indices~\cite{lu2024llava}) and implicit (e.g., relative positions~\cite{huang2025lita}) modeling strategies (Section~\ref{temporal feature subsec}). However, there is no clear consensus on the most effective approach, nor a comprehensive understanding of the trade-offs between these strategies.

Future research should focus on developing unified and expressive temporal encoding mechanisms that effectively capture both fine-grained event details and long-term dependencies. This entails exploring novel representations that embed temporal attributes such as duration, order, and causal relationships directly into the latent space of MLLMs. Furthermore, integrating temporal reasoning tasks into pretraining, such as sequence prediction, duration estimation, and multi-step causal reasoning, could substantially improve grounding precision and robustness.

\textbf{Multimodal Integration}: While most current VTG-MLLMs rely on visual and textual inputs, real-world videos are inherently multimodal, often containing rich audio signals that provide complementary temporal cues. For example, sounds like speech, footsteps, and background noise can serve as precise temporal markers, enhancing grounding accuracy, particularly in visually ambiguous scenes~\cite{viertola2025temporally}, \cite{yariv2024diverse}. However, effectively integrating these additional modalities presents several challenges, including the need for precise temporal alignment, the scarcity of large-scale, multimodal datasets with accurate temporal annotations, and the increased computational complexity.

To address these issues, future work should focus on developing multimodal fusion architectures that can jointly model audio-visual signals with high temporal precision. Potential approaches include leveraging self-supervised learning to pretrain multimodal encoders, designing cross-modal attention mechanisms that dynamically weight different input streams, and creating synthetic multimodal training data to reduce the reliance on expensive manual annotation. Additionally, integrating pre-trained audio-language models, such as Whisper~\cite{radford2023robust}, with visual grounding pipelines could further enhance temporal reasoning across complex, multi-channel video content.

\section{Conclusion}
\label{conclusion}
In this survey, we provide a comprehensive review of VTG-MLLMs, a critical area in fine-grained video understanding. MLLMs have significantly transformed VTG by introducing advanced reasoning and cross-modal alignment capabilities that surpass traditional task-specific methods. We systematically categorize and analyze current approaches, distinguishing between the functional roles of MLLMs as facilitators and executors. Additionally, we examine diverse training paradigms, including pretraining, fine-tuning, and training-free methods, alongside cutting-edge video feature processing techniques, such as visual feature compression and explicit versus implicit temporal modeling. To support ongoing research, we provide a structured overview of widely used datasets and benchmark comparisons, highlighting key performance trends across different VTG-MLLM architectures. Finally, we identify current challenges and propose future research directions to address limitations in training efficiency, temporal representation, and multimodal integration.

\ifCLASSOPTIONcaptionsoff
  \newpage
\fi

{\footnotesize
\bibliographystyle{IEEEtranN}
\bibliography{main}
}

% \input{X_suppl}

% \clearpage
% \input{appendix}

\end{document}